%% file: main_arxiv.tex
\PassOptionsToPackage{table}{xcolor}
\documentclass{template}
\newcommand{\templateoption}{option3}
\usepackage[T1]{fontenc}
\usepackage{lmodern}
\usepackage{xspace}
\usepackage{booktabs}
\usepackage{makecell}
\usepackage{graphicx,amsmath,amssymb,hyperref}
\usepackage[authoryear,round]{natbib}
\usepackage{xcolor}
\usepackage{wrapfig}
\usepackage[percent]{overpic}
\usepackage{url}
\usepackage{siunitx}
\usepackage{verbatim}
\usepackage{makecell}
\usepackage{enumitem}

\hypersetup{
  colorlinks=true,
  urlcolor=magenta,
  linkcolor=black,
  citecolor=link,
  filecolor=black,
  pdfborder={0 0 0}
}
\definecolor{link}{HTML}{0063BE}
\providecommand{\equationname}{Equation}
\providecommand{\sectionname}{Section}
\providecommand{\appendixname}{Appendix}

\newcommand{\figref}[1]{%
  \figurename~\hyperref[#1]{\textcolor{link}{\ref*{#1}}}%
}
\newcommand{\tabref}[1]{%
  \tablename~\hyperref[#1]{\textcolor{link}{\ref*{#1}}}%
}
\newcommand{\eqrefc}[1]{%
  \equationname~\hyperref[#1]{\textcolor{link}{\ref*{#1}}}%
}
\newcommand{\secref}[1]{%
  \sectionname~\hyperref[#1]{\textcolor{link}{\ref*{#1}}}%
}
\newcommand{\appendixref}[1]{%
  \appendixname~\hyperref[#1]{\textcolor{link}{\ref*{#1}}}%
}

\renewcommand{\paragraph}[1]{\textbf{#1.}}
\newcolumntype{x}[1]{>{\centering\arraybackslash}p{#1pt}}
\newcolumntype{y}[1]{>{\raggedright\arraybackslash}p{#1pt}}
\newcolumntype{z}[1]{>{\raggedleft\arraybackslash}p{#1pt}}
\setlist{nosep}

\usepackage{subcaption}
\usepackage{float}

\usepackage{multirow}
\usepackage[normalem]{ulem}
\usepackage{xcolor}
\newlength\savewidth
\title{\centering \fontsize{18}{24}\selectfont{
\textsc{StateM}: Reaching 95.3\% Raw Accuracy, or a \$15 Frontier
Run, on Terminal-Bench 2.1 via Harness Scaling
}
}

\usepackage{makecell}
\usepackage{array}
\newcommand{\captiontitle}[1]{\textbf{#1.}}
\definecolor{wbred}{HTML}{FF8988}
\definecolor{wborange}{HTML}{FECC81}
\definecolor{wbyellow}{HTML}{F5DC51}
\definecolor{wbblue}{HTML}{6098FF}
\definecolor{wbgreen}{HTML}{77B25D}
\definecolor{wbpurple}{HTML}{B28CFF}
\definecolor{wbgray}{HTML}{9AA0A6}
\definecolor{red}{HTML}{FF8988}
\definecolor{orange}{HTML}{FECC81}
\definecolor{yellow}{HTML}{F5DC51}
\definecolor{blue}{HTML}{6098FF}
\definecolor{green}{HTML}{77B25D}
\definecolor{purple}{HTML}{B28CFF}
\definecolor{gray}{HTML}{9AA0A6}

\author{
    \vspace{.2cm}
    \parbox{\textwidth}{\centering
        \Authfont
        Ziheng Qin \hspace{.1em} 
        Yaxin Lu \hspace{.1em} 
        Zhangyang ``Atlas'' Wang \hspace{.1em} 
        Kai Wang
    }
    \\
    \vspace{.3cm}
    {\normalfont\fontsize{11}{15}\selectfont
    {\textsuperscript{}Somewhere on the Earth}\hspace{.1cm}}
    \footnote{This work was conducted in the authors' \textit{personal time} and does NOT reflect the views of any affiliated organization. Ziheng and Yaxin are core contributors (\textit{i.e.,} equally-first authors).}

}

\makeatletter
\renewcommand{\abscontent}{}%
\makeatother

\newenvironment{abstractblock}{%
  {\centering\large\bfseries\sffamily Abstract\par}
  \vspace{0.2em}
  \begin{list}{}{%
      \setlength{\leftmargin}{2em}
      \setlength{\rightmargin}{2em}
      \setlength{\topsep}{0pt}
      \setlength{\parsep}{0pt}
  }
  \item[]
}{%
  \end{list}
  \par\normalfont\vspace{1em}
}

\makeatletter
\fancypagestyle{firststyle}{
  \fancyhead[L]{}
  \fancyhead[C]{}
  \fancyhead[R]{}
  \fancyfoot[L]{\footerfont \the\correspondingauthor}
  \fancyfoot[R]{}
}
\makeatother
\makeatletter
\DeclareRobustCommand\bfseries{%
  \not@math@alphabet\bfseries\mathbf
  \fontseries\bfdefault\selectfont
  \sffamily
}
\makeatother
\DeclareTextFontCommand{\textbf}{\bfseries\sffamily}

\begin{document}

\begingroup
\makeatletter
\let\raggedright\centering
\makeatother

\maketitle
\endgroup

\begingroup

\addtocounter{footnote}{-1}
\endgroup
\newcommand{\abstractcontent}{%
Long-horizon agents can fail even when their underlying models can solve the
constituent steps. They may lose track of mutable state, fail to reactivate
lessons from earlier executions, skip known procedures, or stop prematurely. We bet on \emph{harness scaling} to improve the
execution system around an agent without changing its model weights. We
introduce \textsc{StateM}, an agent-native runtime that organizes execution
around durable states, phase-local context, checked transitions, recoverable
runbooks, and versioned procedural practices that agents and users can inspect together.

On Terminal-Bench 2.1, GPT-5.5 xhigh with \textsc{StateM} reaches 92.1\%
accuracy, compared with an 83.1\% reference and surpassing GPT-5.6 Sol Ultra (91.9\%). The same runbook transfers to
GPT-5.6 without modification. With GPT-5.6 Sol xhigh, the system records
\textbf{95.3\% raw accuracy} across 445 public-submission trials and succeeds at least once on each of the 89 tasks. The frozen profile also raises GPT-5.6 Luna from 76.7\% to 85.4\%, numerically above the 84.9\% Sol xhigh reference.

Next, starting from the same runtime, runbook structure, and golden
rules, it takes less than \$38 of adaptation cost to raise DeepSeek-V4 Flash from 82.7\% to
88.1\% on the full benchmark under standard timeouts. It reaches 89.1\% on a
disclosed 88-task common core. Extending the timeout only for the remaining
latency-sensitive task yields a descriptive full-suite aggregate matching the
reported 88.8\% GPT-5.6 Sol max result. The complete DeepSeek final-score
evidence costs approximately \textbf{\$15 in API usage}, compared with the \$574.68
reported for the GPT reference. All recorded DeepSeek adaptation and
evaluation API expenditure totals \$52.22.

On BusinessBench, we develop family-specific runbooks on separate development sets using the same harness-development principles. The first frozen held-out evaluation yields smaller aggregate gains of 0.55 macro and 1.34 micro percentage points, while two mechanism-matched task families improve by 10.04 points. These results suggest that concrete runbook rules generalize when tasks share the relevant execution structure, while the methodology for identifying and enforcing such controls remains applicable across more heterogeneous workflows. \textsc{StateM} converts selected generalizable postmortem findings into persistent, executable preconditions and practices. It thereby makes explicit and enforceable what the agent should know, what it should do, and which learned practices should be reactivated in future runs — all through stateful controls. Core code and several runtime cases is open sourced at \href{https://github.com/henryqin1997/statem}{github}.
}

\newcommand{\linkstablestyleone}{%
    \begin{center}
        \small
        \renewcommand{\arraystretch}{1.2}
        \begin{tabular}{rll}
            \worldwideweb & \textbf{Website} & \url{https://www.test.com/}\\
            \github & \textbf{Code} & \url{https://github.com/test}\\
            \hf & \textbf{Data} & \url{https://huggingface.com}
        \end{tabular}
    \end{center}%
}

\newcommand{\linkstablestyletwo}{%
    \noindent\small
    \textbf{Website:} \url{https://www.test.com/}\\
    \textbf{Code:}    \url{https://github.com/test}\\
    \textbf{Data:}    \url{https://huggingface.com}
}

\makeatletter
\ifthenelse{\equal{\templateoption}{option1}}{
    \vspace{-0.1cm}
    
    \teaserfigure{!ht}{\textwidth}

    \vspace{0.3cm}

    \begin{abstractblock}
    \abstractcontent
    \end{abstractblock}

    \newpage
}{
    \ifthenelse{\equal{\templateoption}{option2}}{
        \vspace{-0.3cm}
        
        \begin{abstractblock}
        \abstractcontent
        \end{abstractblock}

        \vspace{0.0cm}

        \teaserfigure{!bh}{\textwidth}

        \newpage
    }{
        \begin{abstractblock}
        \abstractcontent
        \end{abstractblock}

        \vspace{0.5cm}

        \vspace{-0.5cm}
    }
}
\makeatother

\section{Introduction}
\label{sec:introduction}

\input{aw_intro}

\input{aw_method}

\input{aw_exp}

\section{Discussion: Runtime Design for Harness Scaling}
\textbf{Harness scaling requires a runtime that is both agent-native and enforceable.}
The experiments above highlight a practical requirement that is easy to miss when viewing a harness only as a fixed workflow: the runtime itself must be cheap to modify. Conventional state-machine and graph-based agent runtimes provide explicit orchestration and strong control, but typically externalize substantial reasoning into manually constructed nodes, edges, and handlers. As the control policy evolves, modifying this structure becomes an engineering task of its own, making rapid failure-driven iteration increasingly heavy. At the other extreme, agent-native systems such as Codex and Claude Code allow the model to plan and revise its behavior fluidly inside the agent loop, but much of this control remains model-mediated: plans, checklists, and self-review can guide behavior without providing the same persistent, auditable, and externally enforceable transition semantics. \textsc{StateM} is designed to combine these two properties. The agent continues to reason and act in its native interaction loop, while states, hooks, transition conditions, and practices provide a control surface that both agents can modify and humans can inspect and enforce. This makes the control layer itself amenable to rapid iteration---a prerequisite for \emph{harness scaling}, where repeated execution experience is used to continuously improve the runtime rather than only the model or prompt.

\textbf{Beyond single-agent control.}
This work focuses on controlling the execution of a single autonomous agent, but the same separation of state, context, and permissions also motivates multi-agent runtimes. In long-running work, a single agent can become overconfident in an early task interpretation and continue reasoning inside a context shaped by its own previous decisions. Separating agents by role can provide different context views and independent decision boundaries, while role-specific read/write permissions can restrict which artifacts, state, or actions each agent may modify. Such isolation may be useful for review, verification, delegation, and recovery from decisions that would otherwise become self-reinforcing within one context. We are exploring these extensions, but they introduce a separate set of coordination, state-sharing, and permission-design questions. We therefore restrict the present work to single-agent control flows and leave role-isolated multi-agent runtimes to future work.

\section{Conclusion}

This work studies harness scaling as a capability axis complementary to
model scaling. We introduce \textsc{StateM}, an agent-native control
layer that externalizes mutable execution state, refreshes phase-local
context, and checks consequential transitions while preserving the
unified reasoning loop of a general-purpose CLI agent. The agent-native and enforceable features of \textsc{StateM} make harness scaling feasible in efficiency and cost.

The experiments expose two different frontiers. At the quality
frontier, GPT-5.6 Sol xhigh with a runbook frozen from GPT-5.5 records
95.3\% raw accuracy on Terminal-Bench 2.1, or 424 successful trials out
of 445, with at least one success on every task. This remains a public,
pre-adjudication submission rather than a finalized leaderboard result.
At the cost frontier, an adapted DeepSeek-V4 Flash system reaches
88.09\% on the full benchmark under standard timeouts and 89.09\% on
the disclosed 88-task common core. Its final evidence costs
approximately \$15 in DeepSeek API usage, while the complete recorded
DeepSeek API expenditure for adaptation and evaluation is \$52.22.

The transfer results show that procedural control is hierarchical rather
than universal. Concrete practices transfer unchanged across the
GPT-5.5 and GPT-5.6 generations, but do not transfer unchanged to
DeepSeek-V4 Flash. Across providers, the runtime and failure-driven
development methodology remain reusable, while model-dependent
practices must be adapted. BusinessBench yields the corresponding
task-side lesson: frozen aggregate generalization is positive but
modest, and the largest gains occur where control is sparse and matched
to an explicit execution boundary.

More broadly, our results distinguish three ways in which an otherwise
capable model can fail. Relevant knowledge may be unavailable at the
decision point; a lesson from prior experience may not be retained or
activated; or an active procedure may not be completed. \textsc{StateM}
addresses these operational gaps through state-local context, versioned
procedural practices, and checked transitions, respectively.

\textsc{StateM} does not make every model universally stronger, and one
runbook does not fit every workflow. Its value is to make execution
failures inspectable and to convert selected lessons into reusable
control. The central scaling problem is therefore not how many rules a
harness remembers, but which lessons should persist, where they should
intervene, and whether they survive a change of model or task.
Model scaling expands what an agent can do; harness scaling helps the
system remember what experience has taught it to do without training, and ensures that it actually does so before proceeding.

\section{Acknowledge}
We thank Zekai Li and Mengxuan Wu for discussions and feedback on this work.

{
    \small
    \bibliographystyle{ieeenat_fullname}
    \bibliography{main}
}

\appendix
\section{Runbook example}
\label{app:runbook_example}
We here share an example of coding agent runbook:
\begin{verbatim}
name: coding-agent-loop
initial: start

nodes:
  start:
    prompt: |
      Initialize the run context from durable project files. Read the task description,
      progress.md, StateM history, architecture notes, and any project rules
      before planning. It is fine to inspect the full runbook with
      `StateM state`; use `cur`, `next`, and `goto` to execute the current step
      with discipline. Do not use `/clear` as part of the normal loop; use the
      session_refresh node and safe compaction when a later cycle needs cleaner
      context.
    in_hook:
      type: message
      text: "Load the task brief, progress.md, StateM history, and any 
      architecture notes before planning."
    before_transfer:
      type: checklist
      items:
        - Task brief was read
        - Relevant files were inspected
        - Constraints were noted

  plan:
    prompt: |
      Generate an implementation plan from the task context, current progress,
      architecture notes, and project rules.
    before_transfer:
      - type: manual
        prompt: "Plan has been reviewed and is concrete enough to execute."
      - type: predicate
        path: ../progress.md
        exists: true
        non_empty: true

  execute:
    prompt: |
      Execute the plan while keeping edits scoped. Update progress before
      leaving this state. If the concrete task or implementation approach
      creates state-specific review needs, register current-entry dynamic
      checks before leaving execute.
    in_hook:
      type: message
      text: "Use `StateM dynamic path` and `StateM dynamic write` to register 
      task-specific current-entry checks when the implementation needs them."
    before_transfer:
      type: checklist
      items:
        - Implementation matches the plan
        - Tests or verification were run
        - No unrelated files were changed
    out_hook:
      type: message
      text: "Update progress.md before moving on."

  review:
    prompt: |
      Review the implementation before handing work back to the user. Compare
      the code against the plan, project docs, constraints, and golden rules.
      If the implementation differs from the plan, decide whether the change is
      justified and make sure the new consideration is recorded in progress.md
      or the final handoff. Prefer returning to execute when the review finds a
      fixable gap.
    before_transfer:
      type: checklist
      items:
        - Implementation follows the task and current plan, or deviations are justified
        - Relevant docs, architecture notes, and local conventions were followed
        - Constraints and golden rules were checked explicitly
        - Edited file scope is appropriate and unrelated changes were not reverted
        - Tests, lint, or equivalent verification were run, or blockers are recorded
        - Remaining risks and follow-ups are clear enough for the user handoff
    out_hook:
      type: message
      text: "Record review notes, verification results, and any remaining risks
      before leaving review."

  handoff:
    prompt: |
      Hand the run back to the user. Summarize what changed, verification run,
      current state, known risks, and the recommended next command if work will
      continue later.
    in_hook:
      type: message
      text: "Before replying to the user, read progress.md and StateM history 
      so the handoff is grounded in durable state."

  session_refresh:
    prompt: |
      Use this node only after a complete review pass when another loop will
      continue and the context contains stale attempts or noisy intermediate
      output. Generate a safe compaction instruction with
      `StateM compact-prompt --run-id <id>`, compact through the host UI, then
      restore attention with `StateM cur` and `StateM history`.
    before_transfer:
      type: checklist
      items:
        - progress.md and StateM history contain the durable facts needed after compaction
        - Safe compaction was applied, or the context was judged clean enough to skip it
        - Current state and next transition were rechecked after compaction

edges:
  - from: start
    to: plan
    condition: "Initial context is loaded."

  - from: plan
    to: execute
    condition: "Plan is ready for execution."

  - from: execute
    to: review
    condition: "Implementation is ready for review."

  - from: review
    to: execute
    condition: "Review found changes that need another execution pass."

  - from: review
    to: handoff
    condition: "Review passed, or remaining issues and blockers are documented for the user."

  - from: review
    to: session_refresh
    condition: "Review passed and another loop will continue in a cleaner session."

  - from: session_refresh
    to: plan
    condition: "Session context is compacted or intentionally kept, and state was restored."

\end{verbatim}

\section{Our Experiment Equipment and Budget}
The main computation resouce, is personal Codex pro plan and Macbook Pro 2025 (M4 chip). Thanks again to Codex's pro plan so we can have chance to do such an individual research work within 200\$ budget, and the actual usage is less than 125\$ of it. For the reported formal run submission, we use AWS m7i.4xlarge, because daytona would encounter sandbox/verifier/network/pytorch-install timeout more frequently, and personal Macbook Pro 2025 (M4 chip) is not compatible with tune-mjcf kernel requirement. On a personal Macbook Pro 2025, GPT-5.5 xhigh can reaches around 91\% pass rate on Terminal Bench 2.1.

\section{Additional Figures and Tables}

\begin{figure*}[h]
    \centering
    \includegraphics[width=\linewidth]{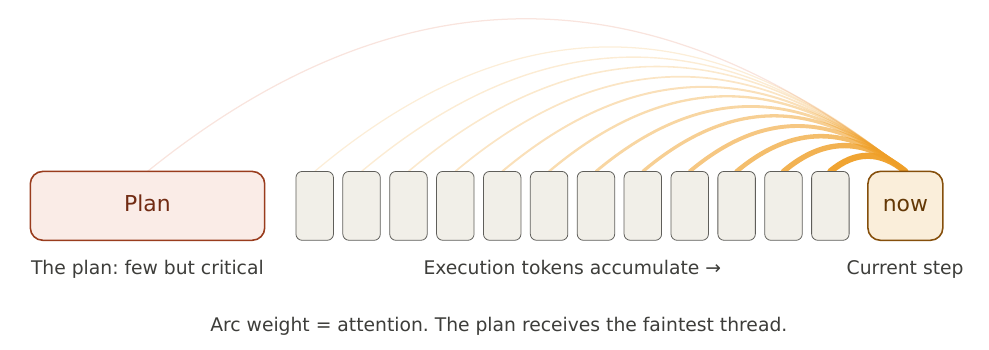}
    \caption{Attention flaw for mixed plan and execution tokens. Plan tokens are at higher abstraction level, which is much fewer but critical. They are diluted by execution tokens in a long run.}
    \label{fig:attention_dis}
\end{figure*}

\begin{figure*}[tp]
    \centering
    \begin{subfigure}[t]{0.48\linewidth}
        \centering
        \includegraphics[width=\linewidth]{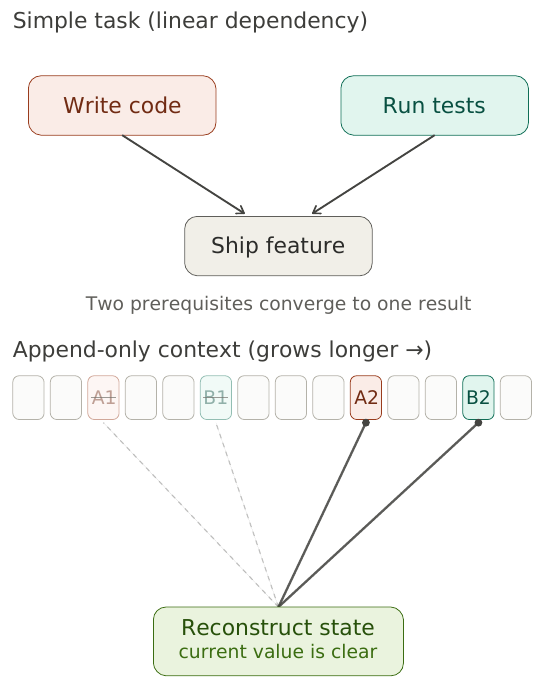}
        \caption{Simple task with linear dependencies: few state updates are appended, so the latest version of each item is easy to locate and the current state recovered without ambiguity.}
        \label{fig:state_simple}
    \end{subfigure}\hfill
    \begin{subfigure}[t]{0.48\linewidth}
        \centering
        \includegraphics[width=\linewidth]{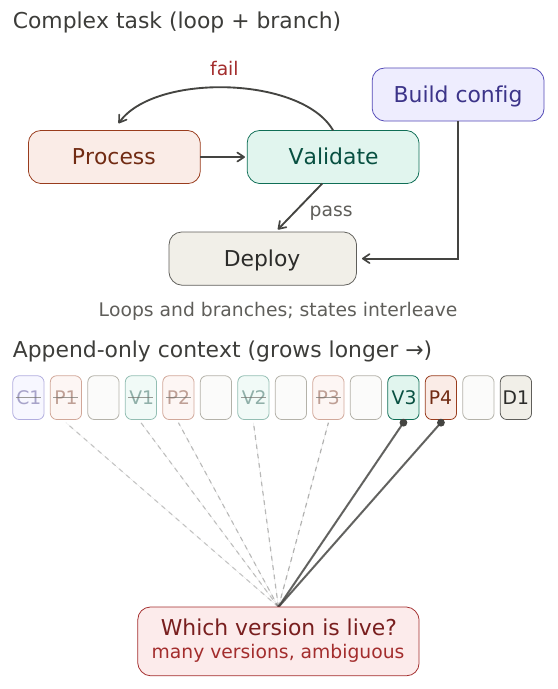}
        \caption{Task with loops and branches: repeated iterations append many interleaved versions of the same item, making the current state ambiguous.}
        \label{fig:state_complex}
    \end{subfigure}
    \caption{\captiontitle{State maintenance degrades with task complexity} Under an append-only context with no in-place editing, each update is appended rather than written back. For a simple task (left) the latest value is still recoverable; for a task with loops and branches (right) the interleaved versions make the current state hard to reconstruct.}
    \label{fig:state_maintenance}
\end{figure*}

\newpage

\begin{table*}[t]
\centering
\small
\setlength{\tabcolsep}{5pt}
\renewcommand{\arraystretch}{1.18}
\begin{tabular}{p{3.25cm} p{4.05cm} p{3.30cm} p{3.40cm}}
\toprule
System & Result source / profile status & Evaluation scope & Score / five-trial coverage \\
\midrule
GPT-5.5 xhigh + Codex & Published reference & 89 tasks, 445 trials & 83.1\%; coverage not reported \\
GPT-5.5 xhigh + StateM & Our developed profile & 89 tasks, 445 trials & \textbf{92.1\%}; 88/89 coverage \\
\midrule
GPT-5.6 Sol xhigh + Codex & Published reference & 89 tasks, 445 trials & 84.9\%; coverage not reported \\
GPT-5.6 Sol xhigh + StateM & Frozen GPT profile; public submission & 89 tasks, 445 trials & \textbf{95.28\% raw}; 89/89 coverage \\
GPT-5.6 Luna + Codex & Published reference & 89 tasks, 445 trials & 76.7\%; coverage not reported \\
GPT-5.6 Luna + StateM & Frozen GPT profile & 89 tasks, 445 trials & \textbf{85.4\%}; coverage not reported \\
\midrule
DeepSeek-V4-Flash & Our baseline & 89 tasks, standard timeout & 82.7\% \\
DeepSeek-V4-Flash + StateM & Frozen GPT profile & 89 tasks, standard timeout & 82.0\% \\
DeepSeek-V4-Flash + StateM & Adapted profile & 89 tasks, standard timeout & \textbf{88.09\%} (392/445) \\
DeepSeek-V4-Flash + StateM & Adapted profile & 88-task common core & \textbf{89.09\%} (392/440) \\
DeepSeek-V4-Flash + StateM & Adapted profile; descriptive aggregate & 89 tasks, one extended-timeout task & \textbf{88.76\%} (395/445) \\
\bottomrule
\end{tabular}
\caption{\textbf{Terminal-Bench 2.1 system-level results and transfer regimes} Reference rows reproduce the stated comparator result; StateM rows report our runs. The descriptive DeepSeek aggregate replaces only the five \texttt{gpt2-codegolf} trials with the disclosed extended-timeout evaluation}
\label{tab:tb21_system_results}
\end{table*}

\begin{table*}[h]
\centering
\footnotesize
\setlength{\tabcolsep}{4.5pt}
\renewcommand{\arraystretch}{1.20}
\begin{tabularx}{\textwidth}{
  >{\raggedright\arraybackslash}p{2.55cm}
  >{\raggedright\arraybackslash}X
  >{\raggedright\arraybackslash}X
  >{\raggedright\arraybackslash}p{2.45cm}}
\toprule
Abstract failure mode
& Generalized StateM control
& Enforcement and evidence
& Main applications \\
\midrule

Phase and side-effect leakage
& Separate read-only discovery and planning from execution. File edits and
external mutations are permitted only after the task-derived plan is ready.
& Workspace snapshots and read-only locks enforce the phase boundary;
subsequent scope checks compare against the active task workspace.
& RefactorBench, WebTest, side-effect workflows \\

Arithmetic and dependency drift
& Route consequential arithmetic through exact-decimal tools, separate numeric
inputs from narrative metadata, and reconcile derived values before acting.
& Compact calculation templates record task-visible operands, operations, units,
rounding, and independently checked outputs.
& Budget approval, payroll, machine operations \\

Missing, duplicated, or partial effects
& Use a task-derived effect manifest followed by execute and fresh-read
reconciliation. Every mandatory effect must succeed exactly once before handoff.
& Structured receipts compare declared, attempted, succeeded, failed, and
freshly observed effects; optional effects remain explicitly distinguishable.
& Budget approval, WooCommerce, WebArena \\

Cross-destination entity mismatch
& Freeze one global task-derived entity set, then derive each destination's
write delta from its observed state and selected write mode.
& Batch closure checks submitted, succeeded, failed, and observed entity IDs,
cardinality, deduplication, and destination-required fields.
& WooCommerce and other multi-sink workflows \\

Bulk-data distortion and context bloat
& Keep tables and repeated records at the tool or file boundary instead of
repeatedly transporting them through the dialogue context.
& Batch tools consume structured inputs and return compact counts, aggregates,
errors, and receipts rather than full payloads.
& Machine operations, payroll, commerce workflows \\

Transient versus deterministic failure
& Retry only failures classified as transient, with a bounded retry budget;
do not retry contract, validation, or deterministic task failures unchanged.
& Each retry uses fresh state where necessary and is followed by a public
fresh-read check. Terminal errors remain explicit incomplete outcomes.
& All tool-using families and benchmark runner \\

Service and session persistence
& Ensure required services survive the agent command session and verify them
through a fresh client or authenticated round trip.
& Readiness depends on observable behavior, stable process ownership, and
post-action state rather than fixed sleeps or shell-local process existence.
& WebArena, WebTest, machine operations \\

Visible-contract and scope drift
& Prioritize visible-contract fidelity and required reference closure;
compatibility is preserved only with visible evidence, and minimality is a
tie-breaker rather than the primary objective.
& Read-only contract extraction, task-derived edit scope, focused public tests,
and a final repository-wide stale-reference scan.
& RefactorBench and repository maintenance \\

Ambiguous API migration
& Prefer named keyword arguments when the visible contract names a parameter,
and prefer direct symbol imports when consistent with the requested migration.
& These are soft defaults. Positional-only APIs, cycles, lazy or optional
imports, monkeypatching, module identity, and repository conventions override
them when supported by public evidence.
& RefactorBench and general code changes \\

Incorrect test or tool boundary
& Keep inspection physically read-only and require authored tests to use the
configured evaluator-free public runner rather than an incompatible host tool.
& The workspace unlocks only on execute; host test entry points are shadowed,
while public-runner results provide the verification receipt.
& WebTest \\

Verification overhead and repeated gating
& Generate templates and receipts at tool boundaries, advance state from
validated tool events, and compress multiple checks into one evidence-bearing
boundary.
& Normal paths avoid repeated StateM round trips. A read-only reviewer runs only
for elevated visible-contract risk and sufficient remaining time; unknown
findings never block.
& All families; reviewer used for RefactorBench \\

\bottomrule
\end{tabularx}
\caption{\textbf{Generalized Businessbench workflow controls distilled into StateM.}
Each control targets an abstract and reusable failure mode rather than a
benchmark-specific answer. Configuration is derived only from the visible task,
the active workspace, public documentation, and observable tool results.
Hard gates are reserved for evidence-bearing invariants such as mandatory-effect
completion, edit boundaries, and batch closure; preferences that may conflict
with legitimate repository designs remain soft and overridable.}
\label{tab:generalized-statem-controls}
\end{table*}

\end{document}

%% file: aw_intro.tex
Long-horizon agents often fail in a revealing way: the underlying model appears
capable of solving each local step, yet the complete run still fails. The agent
deviates from its plan, loses track of mutable task state, skips a necessary
check, repeats an unproductive action, or stops before the requested deliverable
is verifiably complete~\citep{yao2024tau,barres2025tau2,deng2025swebenchproaiagents,zhu2026agentaging,liu2026evaluatingplancomplianceautonomous}.
These failures matter because real workflows and agent benchmarks evaluate the
completed job, not whether the model knew how to perform most of its constituent steps.

The dominant response to such failures is to improve the model: scale pretraining, add post-training data, increase test-time reasoning, or introduce
additional agents. We study an orthogonal question:

\begin{quote}
\emph{How much apparent model failure is actually failure of the harness that
maintains state, constrains execution, verifies progress, and recovers from
errors?}
\end{quote}

We call the systematic improvement of this surrounding control layer
\emph{harness scaling}. Harness scaling does not replace model scaling. It asks
whether more of a model's existing capability can be converted into completed,
reliable work by improving the runtime around it. This leads to three
progressively stronger empirical tests: \textit{first}, can a better harness improve a
fixed model without changing its weights? \textit{Second}, can a harness developed with
one model transfer to a newer model without retuning? \textit{Third}, can the resulting
control principles improve performance beyond the benchmark on which they were
developed?

\paragraph{Why external control?}
We motivate harness scaling through two design hypotheses. First,
\emph{control-signal dilution} can arise when a compact plan and its completion
criteria are surrounded by an increasingly long trace of commands,
observations, and repairs. Second, \emph{mutable-state ambiguity} can arise when
completed goals, pending dependencies, failed attempts, and valid next actions
must be reconstructed from an append-only history rather than read from an
authoritative current state. We treat these as operational hypotheses, not as
fundamental claims about transformer attention. Both suggest externalizing
procedural state, refreshing the control information relevant to the current
phase, and checking evidence before important transitions.

\begin{figure*}[t]
    \centering
    \includegraphics[width=\linewidth]
    {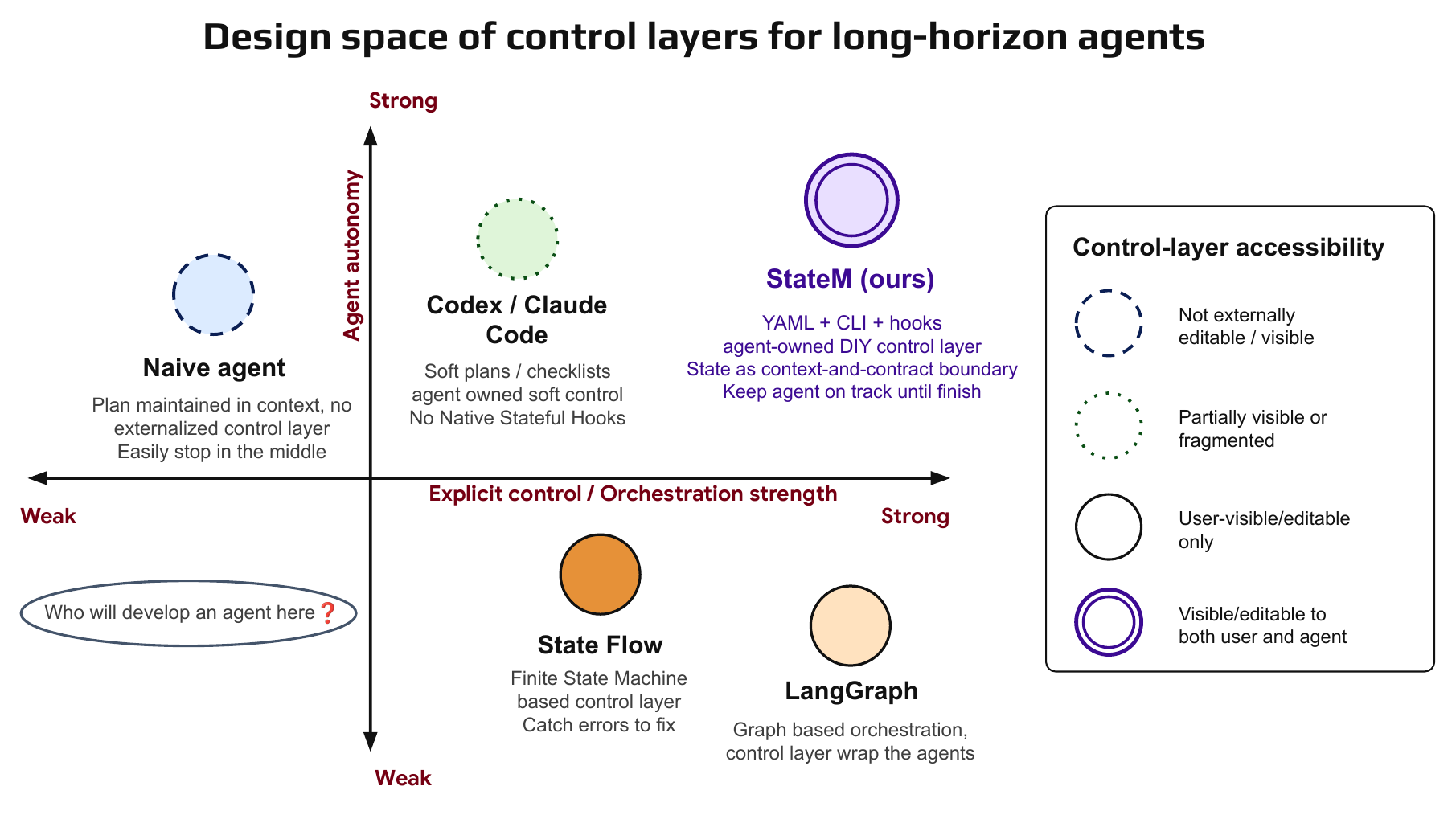}
    \caption{
    \textbf{Conceptual design space for agent control layers.}
    The vertical axis denotes \emph{runtime enforceability}: the extent to which
    the runtime maintains authoritative execution state, constrains valid
    transitions, and can block incomplete handoffs. The horizontal axis denotes
    \emph{agent autonomy}: the latitude retained by the primary agent to choose
    tools, revise procedures, and operate over a broad action space. Marker
    style denotes whether the control artifact is primarily developer-owned,
    agent-editable, or jointly editable by the agent and user. Positions
    represent typical operating abstractions rather than architectural limits
    or empirical performance rankings. StateM targets the upper-right region
    through a shared runbook that combines broad agent autonomy with explicit,
    enforceable state transitions. Evaluation and harness-optimization methods,
    including AgingBench and automated harness adaptation, are discussed in the
    text but are not plotted because they operate on different dimensions.
    }
    \label{fig:design_space}
\end{figure*}

\paragraph{A control-layer design space}
Existing systems expose a tradeoff between runtime enforceability and agent
autonomy. State-machine and graph-based runtimes provide explicit workflow
state, persistence, conditional routing, and recovery
~\citep{wu2024stateflowenhancingllmtasksolving,
wang2024agentailanggraphmodular}, but typically organize execution around a
developer-authored controller. General-purpose CLI agents preserve a broader
reasoning and action space, but their plans, instruction files, memories, and
hooks do not by themselves form a unified, transition-aware control surface.
Figure~\ref{fig:design_space} summarizes these typical operating points. The
distinction is not whether a system has state, but who can inspect, modify, and
operate the stateful control layer during execution.

A complementary line of work studies reliability over even longer timescales.
AgingBench shows that an agent with frozen weights remains a changing system as
its memory is compressed, retrieved, revised, and maintained
~\citep{zhu2026agentaging}. Its typed-state and runtime-control interventions
further indicate that some longitudinal failures require explicit state rather
than additional text context alone. 
Related work and the precise boundaries of our contribution are discussed in
Section~\ref{sec:related_work}.

\paragraph{\textsc{StateM}: an agent-native control layer}
We introduce \textsc{StateM}, a lightweight runtime for long-running CLI
agents. Its control layer takes the form of a human-readable YAML runbook
containing states, valid transitions, state-local instructions, hooks, checks,
and recovery rules. Through ordinary command-line operations, the agent can
inspect its current state, request a transition, examine failed conditions,
review execution history, recover after an interruption, and update permitted
runbook artifacts. A human supervisor can inspect, edit, version, and audit
the same runbook.

\textsc{StateM} follows two principles. First, each state is a
\emph{context-and-contract boundary}. Entering a state refreshes the active
instructions and durable task information for that phase. Leaving the state
requires the agent to satisfy explicit exit conditions. The host runtime can
evaluate executable conditions directly. Conditions that require semantic
judgment are recorded as auditable attestations and may trigger further
review or repair. This distinction matters because an agent's declaration of
completion does not constitute independent verification. Second, the agent and user \emph{share} the control layer. The agent remains
the primary executor and retains a unified reasoning loop, but its progress is
exposed through a control surface that the host runtime can enforce and the
user can revise. \textsc{StateM} is agent-native in a specific sense: the
agent operates the control layer through the same CLI action space it uses to
perform the task. The runtime requires no model modification, fine-tuning, or
access to model internals.

\textsc{StateM} combines three properties that are not usually available in
one system:

\begin{enumerate}
    \item state boundaries provide both refreshed control contexts and
    explicit transition contracts;
    \item a general-purpose agent remains the primary executive instead of
    being decomposed into a fixed collection of node-local model calls; and
    \item the runbook is a shared, machine-operable artifact that agents and
    users can inspect, audit, and revise through the same execution
    environment.
\end{enumerate}

This representation allows the harness itself to be optimized. After a run,
failures can be classified as missing context, invalid transitions, weak
checks, premature handoffs, ineffective recovery, or other control-layer
defects. The working agent or a separate hyper-agent can then propose changes
to state boundaries, prompts, hooks, checks, recovery rules, and practice
activation conditions. Candidate changes are reviewed, tested for
regressions, and incorporated into a new runbook version. The model weights
remain unchanged, while selected procedural knowledge accumulates in the
external control layer.

This process addresses a failure mode that arises across executions rather
than within a single run. An agent may correctly diagnose a failure but
neither retain nor invoke the resulting lesson when the same risk recurs. We
call this the \emph{procedural-memory gap}. It complements an
\emph{epistemic gap}, in which relevant knowledge or an appropriate method is
unavailable at the decision point, and a \emph{procedural-compliance gap}, in
which the correct procedure is active but remains incomplete or not completely followed by the agent. State-local
context, versioned practices, and checked transitions target these three gaps,
respectively. A state records the position of the current run. Procedural
memory records what earlier runs have taught the system to do at that point.

We call the resulting development process \emph{failure-driven harness
optimization}. It is related to recent work on improving smaller models
through stronger or automatically adapted harnesses
\citep{yang2026betterharnessessmallermodels}. \textsc{StateM} focuses on a
machine-operable, agent-native runbook representation. We use this
representation to study which layers of procedural control transfer across
model versions, providers, and task families.

\paragraph{Results and scope}
We organize the experiments around two complementary frontiers, summarized in 
Fig.~\ref{fig:tb21_cost_frontier}. At the quality frontier, GPT-5.5 xhigh with
\textsc{StateM} reaches 92.1\% accuracy on Terminal-Bench 2.1, compared with
the 83.1\% GPT-5.5 reference. Across five trials, the system solves 88 of the
89 tasks at least once. Its 92.1\% mean is also numerically close to the
91.9\% next-generation reference. This comparison illustrates that the
observed system-level harness gain can resemble a model-generation upgrade. We then freeze the runbook developed with GPT-5.5 and apply it to GPT-5.6
without modifying its control logic. With GPT-5.6 Sol xhigh, the combined
system records 95.28\% raw accuracy in a public Terminal-Bench 2.1
submission. This result corresponds to 424 successful trials out of 445, with
at least one successful trial on every task.\footnote{
\textbf{Evaluation status as of August 11, 2026.}
The 95.28\% figure is the raw, pre-adjudication accuracy computed by the
Terminal-Bench 2.1 submission pipeline in PR~\#142:
\url{https://github.com/harbor-framework/terminal-bench-2-1/pull/142}.
The submission passed ten automated checks covering task digests, standard
runtime settings, trial coverage, and trajectory availability. The PR remains
open and has not been merged into the leaderboard. We agree that four
rewarded trajectories identified during review should not count. Scoring
those trials as zero gives $420/445=94.38\%$. Scoring all nine trajectories
currently flagged for possible reward hacking as zero gives
$415/445=93.26\%$. We therefore report 95.28\% only as a raw public
submission score. The submission pipeline reports 1.178 billion tokens and
\$1,062.95 in model cost. Five-trial task coverage does not measure
single-run reliability.
}
The frozen runbook also raises GPT-5.6 Luna from 76.7\% to 85.4\%, a gain of
8.7\%. This result is numerically above the 84.9\% Sol xhigh
reference. 

The cost frontier has a transfer boundary. Starting
from the same runtime, high-level runbook structure, and golden rules, less
than \$38 of DeepSeek API expenditure is sufficient to adapt the concrete
practices. The adapted system reaches $392/445=88.09\%$ on the full 89-task
benchmark under standard timeouts. On the disclosed 88-task common core,
which excludes the latency-sensitive \texttt{gpt2-codegolf} task, it reaches
$392/440=89.09\%$. Extending the timeout only for that task yields a
descriptive full-suite aggregate above the reported 88.8\% GPT-5.6 Sol max
result. The complete DeepSeek final-score evidence costs approximately \$15 in
realized API charges, compared with \$574.68 reported for the GPT-5.6 Sol max
reference. All recorded DeepSeek API expenditure across profile adaptation
and final evaluation is \$52.22. These amounts are approximately
$38.9\times$ and $11.0\times$ lower, respectively. 

We also evaluate task-side generalization on BusinessBench~\citep{yang2026betterharnessessmallermodels} using Codex with
GPT-5.6 Luna. The dataset contains 477 eligible instances across seven task
families. Of these, 405 instances across six families receive a
\textsc{StateM} intervention. In the first frozen one-shot evaluation,
held-out performance changes from 84.67\% to 85.22\% under an equal-family
macro average, and from 84.44\% to 85.78\% under an instance-weighted micro average. Development performance improves from 86.07\% to 91.71\%, while
performance across all treated instances improves from 84.76\% to 88.72\%.
An exploratory two-family slice containing Budget Approval and Machine
Operating improves from 71.91\% to 81.94\%. This pattern suggests that the
largest gains occur when the control profile matches a specific execution
boundary.

\paragraph{Contributions}
This work makes five contributions:

\begin{itemize}
    \item We formulate \emph{harness scaling} as a capability axis
    complementary to model scaling. We distinguish within-model improvement,
    frozen cross-generation transfer, adapted cross-provider transfer, and
    held-out task generalization as separate empirical tests.

    \item We introduce \textsc{StateM}, an agent-native runtime in which
    durable states provide refreshed control contexts and explicit transition
    contracts while preserving the unified reasoning loop of a
    general-purpose CLI agent.

    \item We operationalize three sources of execution failure: epistemic
    gaps, procedural-memory gaps across runs, and procedural-compliance gaps
    within a run. State-local context, versioned practices, and checked
    transitions provide corresponding control points.

    \item We develop a failure-driven optimization loop that converts selected
    execution failures into versioned changes to prompts, hooks, checks,
    recovery rules, state boundaries, and practice activation conditions.
    This process allows procedural knowledge to accumulate outside the model
    weights.

    \item We evaluate the system across GPT-5.5, two GPT-5.6 variants,
    DeepSeek-V4 Flash, Terminal-Bench 2.1 and BusinessBench. The evaluation
    covers raw-score adjudication sensitivity, frozen and adapted transfer,
    task-level regressions, realized API expenditure,
    and explicit disclosure of benchmark adaptation.
\end{itemize}

The results do not suggest that model scaling has ceased to matter or that a
static runbook can dominate every model, task, and benchmark. We instead advocate that runtime semantics and principles for harness development can
transfer broadly, while concrete practices transfer locally or require
inexpensive adaptation. Model capability and execution reliability make
separate contributions to system performance. In the settings studied here,
\textbf{the model appears not to be the (main) bottleneck}.

\section{Related Work}
\label{sec:related_work}

StateM intersects four lines of work: long-horizon planning, stateful agent
orchestration, long-lived agent memory, and automated harness adaptation. Its
novelty does not lie in any one component in isolation. Instead, it lies in
their combination through an agent-native, jointly editable execution
representation.

\paragraph{Planning and long-horizon agent execution}
Planning is a standard response to long-horizon complexity: an agent first
decomposes a task and then executes the resulting procedure. Nevertheless,
current agents frequently depart from their plans, omit necessary steps, or
terminate without satisfying the original completion criteria
~\citep{liu2026evaluatingplancomplianceautonomous}. Plans represented only in
natural language remain advisory unless the runtime maintains progress and
checks whether required conditions have been met.

At the same time, stronger orchestration is not universally beneficial.
Decomposing a procedure into many separately prompted nodes can fragment
reasoning and introduce routing errors. Recent evidence finds settings in which
a strong model given the full procedure in context outperforms a more heavily
orchestrated version of the same procedure
~\citep{dennis2026incontextpromptingobsoletesagent}. StateM is motivated by this
tension. It seeks stronger execution control without decomposing the primary
agent into a sequence of narrowly scoped model calls.

\paragraph{Stateful and graph-based orchestration}
Finite-state workflows and graph-based agent runtimes already provide important
control mechanisms. StateFlow represents problem solving through explicit
states and transitions, demonstrating that state-driven execution can improve
task grounding and failure recovery
~\citep{wu2024stateflowenhancingllmtasksolving}. LangGraph and related durable
workflow systems provide persistent state, conditional edges, checkpointing,
interruption, and human-in-the-loop recovery
~\citep{wang2024agentailanggraphmodular}. StateM does not claim to introduce
these capabilities.

The difference is the default execution abstraction. In a conventional
controller-led graph, a developer specifies the workflow and invokes models or
agents within its nodes. The graph may be highly flexible, but the primary
control artifact remains external to the executing agent. StateM instead keeps
a general-purpose CLI agent as the primary executive and exposes the control
layer within its ordinary tool environment. The agent can inspect its current
state, request transitions, examine failed conditions, and propose permitted
runbook changes without leaving its normal action space.

This distinction should not be read as a strict separation between
architectures. A graph runtime can host a highly autonomous agent, and a StateM
runbook could be implemented on top of a durable workflow engine. Figure
~\ref{fig:design_space} describes typical design choices rather than hard
capability boundaries. The specific contribution of StateM is a lightweight
agent-facing control representation in which state boundaries are both
refreshed control contexts and auditable transition contracts.

\paragraph{CLI agents, soft plans, and user control}
General-purpose CLI agents such as Codex and Claude Code preserve broad autonomy
over tool selection, file manipulation, code execution, and iterative problem
solving. They also expose useful control fragments, including planning prompts,
TODO lists, rule files, memory files, and host-side hooks. These mechanisms make
the agents highly adaptable, but they do not necessarily establish one
authoritative execution state. Plans and checklists are usually soft
natural-language artifacts, while hooks are commonly attached to tool or
lifecycle events rather than to semantically meaningful transitions.

StateM organizes these fragments into a shared runbook containing current state,
valid transitions, state-local instructions, checks, hooks, recovery rules, and
execution history. The user can therefore intervene by editing the same control
artifact that the agent and runtime operate, rather than only by adding another
natural-language correction to the interaction history. This shared ownership
is the main distinction represented by the third visual encoding in
Figure~\ref{fig:design_space}.

\paragraph{Long-horizon agents and agent aging}
Long-horizon execution within one task is related to, but distinct from,
reliability across many sessions. AgingBench frames deployed agents as
time-evolving systems whose effective behavior can change even when model
weights remain frozen~\citep{zhu2026agentaging}. It identifies four longitudinal
failure mechanisms: compression aging, interference aging, revision aging, and
maintenance aging. It further localizes failures to writing, retrieval,
utilization, and lifecycle handling through temporal dependency graphs and
counterfactual diagnostic probes.

AgingBench is particularly relevant to StateM in two ways. First, its revision
aging results show that mutable or derived state can fail even when additional
text memory is available, supporting the need for explicit state
representations. Second, AgingBench studies a typed-state overlay and a
threshold-triggered runtime controller as targeted interventions. These results
connect diagnosis to runtime control, but at a different level from StateM. The
AgingBench interventions maintain selected memory variables or activate
memory-policy changes across sessions. StateM provides a general procedural
runtime for representing phases, transitions, evidence, checks, and recovery
within open-ended agent work.

The two approaches are therefore complementary. AgingBench asks how reliability
changes over an operational lifespan, which mechanism is degrading, and where a
repair should target. StateM asks how an agent should represent and enforce its
current procedural state while completing a long-running task. Combining
lifespan diagnostics with StateM runbook adaptation is a natural direction for
future work, but StateM does not by itself solve compression, retrieval, or
maintenance aging.

\paragraph{Harness adaptation and self-improvement}
A growing body of work shows that agent performance depends strongly on the
runtime harness and that the harness can itself be optimized. Agentic Harness
Engineering uses execution observability to evolve coding-agent tools,
middleware, prompts, and memory from trajectory feedback
~\citep{lin2026agenticharness,gu2026model}. Life-Harness converts recurring failures into
reusable interventions over environment contracts, procedural skills, action
realization, and trajectory regulation, then freezes the resulting harness for
held-out evaluation~\citep{xu2026lifeharness}. Self-Harness lets an agent mine
its weaknesses, propose harness modifications, and promote them through
regression-aware validation~\citep{zhang2026selfharness}. Better Harnesses,
Smaller Models studies automated harness adaptation as a way for lower-cost
models to recover much of the performance of larger models
~\citep{yang2026betterharnessessmallermodels}.

These works make two points clear. First, model identity alone is not a
sufficient description of an agent system. Second, neither harness adaptation
nor cross-model harness transfer is, by itself, a novel claim. StateM differs
primarily in the artifact being optimized and operated. Its harness is an
explicit state-machine runbook whose states, transition conditions, hooks,
repair paths, and history are directly available to the executing agent and
human supervisor. The same representation serves three roles: runtime control,
an audit surface, and the search space for failure-driven improvement.

StateM is thus closest to prior harness-evolution methods in its hyper-agent
loop, but its central contribution is the execution substrate that this loop
modifies. Unlike prompt-only optimization, StateM can alter transition-time
checks, recovery behavior, and state-local context. Unlike a fully external
workflow graph, the resulting runbook remains inside the primary agent's normal
tool environment. Unlike longitudinal diagnostic benchmarks, it acts directly
on within-run procedural execution.

\paragraph{Positioning StateM}
The quadrant in Figure~\ref{fig:design_space} summarizes the resulting design
claim. Soft plans and conventional CLI control fragments preserve autonomy but
offer limited transition-level enforcement. Developer-authored workflow graphs
offer strong control but typically place the agent within an externally
specified execution structure. StateM targets the less explored combination of
broad agent autonomy, enforceable state transitions, and a control artifact
jointly visible and editable to the agent and user. This positioning does not imply that one point in the quadrant is universally
superior. Stable, repetitive workflows may be best served by a fixed graph,
while short or exploratory tasks may need little control beyond an
instruction file. StateM is intended for the intermediate regime: tasks that
are open-ended enough to require a general-purpose agent, but long and
consequential enough that soft plans and self-declared completion are
insufficient.

%% file: aw_method.tex
\section{StateM: Agent-Native Control for Long-Horizon Execution}
\label{sec:statem}

Section~\ref{sec:introduction} identified two operational pressures in
long-horizon execution: the control signal carried by a plan can weaken as the
trace grows, and the current task state can become ambiguous when it must be
reconstructed from an append-only history. Section~\ref{sec:related_work}
further identified a design tension between runtime enforceability and agent
autonomy. StateM addresses this tension by preserving broad agent discretion
\emph{within} phases while making progress \emph{between} phases explicit,
persistent, and checkable.

The design has four requirements. First, the control layer should represent
coarse phases of work without decomposing the agent into a sequence of
micro-actions. Second, it should refresh phase-relevant instructions and durable
progress when a phase begins. Third, it should expose explicit control points
where incomplete work, failed verification, or external blockers can prevent a
handoff. Finally, it should be directly operable by the agent and auditable by
the user through their shared workspace.

StateM realizes these requirements as a YAML-configured state-machine runtime
with a command-line interface. A versioned \emph{runbook} defines states,
transitions, prompts, hooks, conditions, and checks. The agent operates this
runbook through ordinary CLI commands, while the runtime maintains execution
state outside the model context. The control layer is therefore externalized
from the interaction history without being hidden from either the agent or the
user.

\begin{figure*}[t]
    \centering
    \includegraphics[width=0.8\linewidth]
    {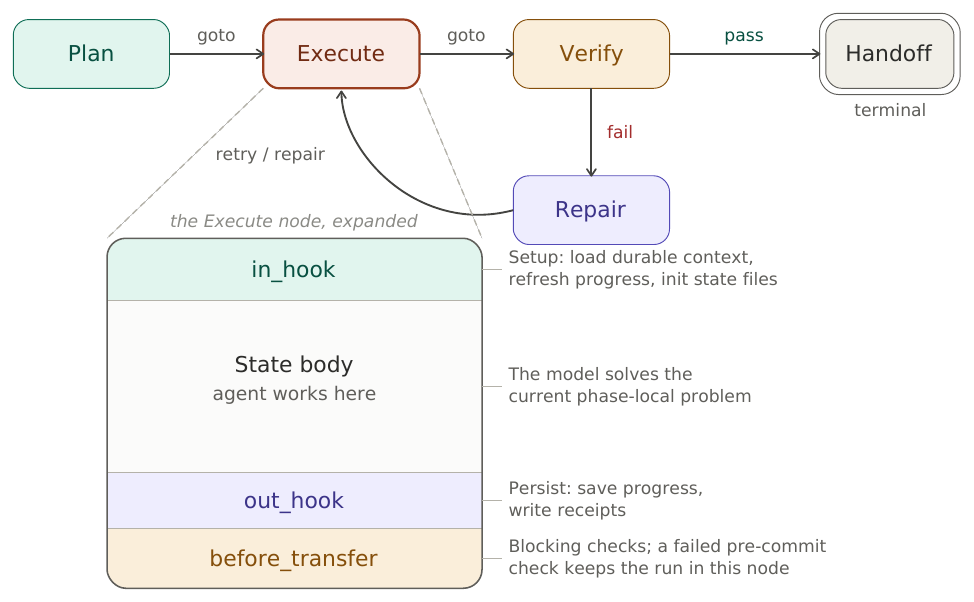}
    \caption{
    \captiontitle{The \textsc{StateM} control surface}
    A runbook is a directed state machine over coarse phases (top).
    Each \texttt{goto} follows a checked and logged transition protocol;
    verification failures can route execution through a configured repair
    state, and successful completion is recorded only at a terminal state.
    Expanding a node (bottom) reveals its phase-local structure:
    an \texttt{in\_hook} for setup and context loading, a state body in which
    the model performs the open-ended work, an \texttt{out\_hook} for
    persistence, and a \texttt{before\_transfer} block containing checks that
    must pass before the transition can commit.
    }
    \label{fig:StateM_design}
\end{figure*}

\subsection{Runtime and Control Profile}
\label{sec:statem_layers}

StateM separates two layers that should not be conflated. The
\emph{runtime} provides the generic mechanisms for state persistence,
transition validation, hook execution, history, and recovery. The
\emph{control profile}, represented by a runbook, specifies the phases,
instructions, checks, and repair policies used for a particular class of work.

This distinction is important for interpreting the experiments. The StateM
runtime is designed to be reusable across agents and workflows, but the content
of a runbook may encode workflow-specific or experience-derived procedural
knowledge. In particular, the Terminal-Bench results in
Section~\ref{sec:experiments} evaluate the combined runtime and an evolved,
benchmark-adapted control profile. They should not be interpreted as isolating
the effect of the state-machine abstraction alone.

The base runbook is a static, shareable, and versionable artifact. Multiple
agents and executions can use the same runbook, while each execution maintains
separate mutable runtime state. Runbooks may be updated between versions, and
permitted run-local additions can be recorded during execution, but such changes
remain distinct from the underlying runtime.

\subsection{States as Context-and-Contract Boundaries}
\label{sec:state_boundary}

The central StateM abstraction is a \emph{phase-level state}. A state represents
a meaningful stage of work rather than a single model call or tool action. A
coding runbook, for example, may include states for planning, implementation,
task-contract checking, self-review, repair, and handoff. The agent remains free
to reason, invoke tools, edit files, and iterate within each state.

A state serves first as a \emph{context boundary}. When the agent enters a
state, StateM exposes the current phase, valid outgoing transitions,
state-local instructions, and relevant durable progress. The
\texttt{in\_hook} performs the configured entry procedure. It may inject a
prompt, run setup code, initialize state-local files, or load a compact progress
record. This creates a fresh control anchor without requiring the agent to
infer its current phase and outstanding obligations entirely from the preceding
terminal trace.

This mechanism does not erase the model's prior context or guarantee lossless
compaction. Rather, it makes the authoritative phase and its local obligations
explicit and recent. It can repair a missing-context problem when the necessary
information is available to the runbook, but it cannot supply task knowledge
that neither the model nor the accessible environment possesses.

A state also serves as a \emph{contract boundary}. The state prompt describes
the work expected within the phase, while exit hooks and checks specify the
conditions for leaving it. The \texttt{out\_hook} can persist progress, update
receipts, or prepare artifacts for the next phase. The
\texttt{before\_transfer} block evaluates the configured exit conditions before
StateM commits a transition.

StateM supports several types of checks with different evidentiary strength:

\begin{itemize}
    \item \texttt{command} and \texttt{predicate} checks are evaluated by the
    host and can provide independently reproducible evidence, subject to the
    correctness of the command or predicate;

    \item \texttt{manual} checks require an explicit user or operator decision;

    \item \texttt{checklist} and \texttt{message} checks require the agent to
    acknowledge specified obligations, but remain forms of structured
    self-attestation; and

    \item \texttt{llm\_review} checks add a separate semantic judgment, but do
    not constitute deterministic verification.
\end{itemize}

This distinction prevents a receipt or model declaration from being treated as
proof merely because it appears in a structured field. StateM makes such claims
visible and auditable; their reliability still depends on the mechanism that
produced and verifies them.

Together, state entry and state exit address different failure modes. Entry
hooks refresh the phase-local control signal and make durable progress
available. Exit checks create intervention points for procedural requirements,
such as running a consumer command, executing tests, recording evidence,
obtaining approval, or updating a receipt. Failed checks remain visible rather
than silently propagating into later phases.

\subsection{Runbooks, Edges, and Transition Verification}
\label{sec:statem_transition}

Formally, a runbook defines
\[
    \mathcal{B}
    =
    \left(
        \mathcal{S},
        s_0,
        \mathcal{S}_{T},
        \mathcal{E},
        \Phi
    \right),
\]
where $\mathcal{S}$ is the set of phase-level states, $s_0$ is the initial
state, $\mathcal{S}_{T}\subseteq\mathcal{S}$ is the set of terminal states,
and $\mathcal{E}\subseteq\mathcal{S}\times\mathcal{S}$ is the set of permitted
transitions. The state specification $\Phi(s)$ contains the phase-local prompt,
entry and exit hooks, transfer checks, and references to state-local artifacts.
A complete example appears in Appendix~\ref{app:runbook_example}.

An edge may additionally contain a guard or transfer hook. These edge-level
conditions select among valid next states using information such as check
results, receipts, previous failures, external readiness, or user-blocking
status. After verification, for example, a runbook may route to
\texttt{repair} when tests fail, to \texttt{handoff} when sufficient evidence
has been recorded, or to \texttt{wait} when an external service or human
approval is required. Repair routing is explicit in the graph rather than an
implicit instruction that the agent must remember.

All phase changes use the core operation \texttt{goto}. When the agent requests
\texttt{goto TARGET}, StateM executes an ordered transition protocol:

\begin{enumerate}
    \item verify that the requested edge from the current state to
    \texttt{TARGET} exists;

    \item evaluate the current state's \texttt{before\_transfer} checks;

    \item run the current state's configured persistence or
    \texttt{out\_hook} operations;

    \item evaluate edge guards and edge-specific transfer hooks;

    \item commit the target state and append the transition event to history
    only if all required pre-commit steps succeed; and

    \item create the target-state entry and execute its \texttt{in\_hook}.
\end{enumerate}

If a required pre-commit check or hook fails, the run remains in the source
state and records the failure, allowing the agent to inspect the unmet
condition, repair the underlying problem, and retry. If the transition
succeeds, StateM creates a new state-entry record and exposes the target
state's instructions and obligations.

We describe this protocol as \emph{checked, logged, and recoverable}, rather
than fully transactional. StateM delays the runtime-state commit until required
pre-commit operations succeed, but it cannot generally roll back arbitrary
external side effects produced by hooks.\footnote{
Hooks execute with the permissions of the surrounding host environment.
Commands that modify external services or non-versioned artifacts should
therefore be idempotent where possible and should record sufficient information
for explicit compensation or retry. StateM does not make untrusted hook code
safe; sandboxing and permission control remain host responsibilities.
}
Recovery refers to the StateM execution record and configured repair procedure,
not ACID rollback of the external world.

\subsection{Per-Run State and Recovery}
\label{sec:statem_recovery}

A runbook describes reusable control logic, while each execution has its own
runtime record. For each run, StateM stores a run identifier, current state,
current state-entry identifier, transition history, hook and check outcomes,
timestamps, and references to state-local evidence files. Separating this
mutable record from the runbook allows one control profile to support multiple
agents or concurrent runs without conflating their progress.

This persistent record also provides a recovery anchor. After a process restart,
context refresh, or model-side compaction, the agent can query StateM for the
current state, prior transitions, unresolved checks, and recorded evidence.
It need not reconstruct the workflow solely from a long terminal transcript.
A restarted agent can therefore resume from an explicit phase and a durable set
of obligations.

Recovery has a precise scope. StateM can restore its recorded control state and
re-execute configured recovery steps. It cannot reconstruct work that was never
persisted, restore hidden model context, or automatically reverse arbitrary
external actions. Reliable recovery consequently depends on placing persistence
hooks at appropriate phase boundaries and designing repeated hooks to tolerate
retry.

A run may be in one of three operational situations. It may be active in a
non-terminal state, paused because an external condition is unresolved, or
complete in a terminal state. Reaching a non-terminal pause is not treated as
successful completion. This distinction allows StateM to separate genuine
handoff from temporary inability to proceed.

\subsection{Shared Control, Runtime Checks, and Stop Hooks}
\label{sec:statem_shared_control}

StateM is agent-native because the executing agent operates the control layer
through the same CLI environment in which it performs the task. The runbook is
a normal workspace artifact rather than hidden controller code. The agent can
inspect its current state, follow configured transitions, examine failures, and
propose runbook changes. The user can read, edit, review, and version the same
artifact. This shared surface preserves broad within-state autonomy while
making cross-state obligations explicit.

Shared ownership does not mean unrestricted self-modification. StateM
distinguishes among the versioned base runbook, run-local additions, and
host-owned constraints. During execution, an agent may register additional
dynamic checks when the runbook permits it. Such checks are useful when the
agent discovers a risk or requirement that was not known when the runbook was
written. The addition is recorded in the run history so that the user can
inspect it and decide whether it should be promoted into a later runbook
version.

Adding a stricter run-local check is different from weakening or deleting an
existing requirement. Changes to user-owned invariants, permissions, or
blocking checks should require a privileged policy decision. YAML editability
alone does not provide this security boundary; deployments must define which
parts of a runbook the agent may modify. In the reported cross-model transfer
experiment, the evaluated base runbook is frozen so that runbook evolution is
not conflated with evaluation-time adaptation.

StateM also supports host-level stop-hook integration for Codex and Claude Code.
When the host receives a request to stop, the hook can inspect the current
StateM status. If the run is in a terminal state or explicitly blocked on an
external dependency, the stop can proceed. Otherwise, the hook returns the
current phase and unmet obligations to the agent and requests continued work.
This mechanism reduces premature termination and supports longer unattended
runs.

A stop hook does not guarantee eventual success or termination. The agent may
remain unable to satisfy a check, exhaust its model or environment budget, or
repeat an ineffective repair. Stop-hook behavior should therefore remain
subject to retry, time, and resource limits. Its benefit is greater execution
persistence and a more explicit stopping condition, not greater reasoning
capability by itself.

\subsection{Scope of the Control Guarantee}
\label{sec:statem_scope}

StateM provides control points, not a correctness oracle. Its guarantees are
strongest when a transition condition can be checked independently by a
command, predicate, or human decision. Semantic self-review and agent-authored
receipts remain fallible. Similarly, StateM can prevent a configured requirement
from being silently skipped, but it cannot detect a missing requirement that
neither the runbook nor the agent introduces.

The framework should therefore be understood as an execution substrate with
three roles. It is a runtime that maintains procedural state and controls
transitions; an audit surface that exposes progress, evidence, and failure; and
an optimization target through which experience-derived procedural knowledge
can be preserved outside model weights. The empirical gains in the following
section arise from the combined runtime and control profile. The next section
examines how that profile was developed, frozen, transferred across model
versions, and evaluated.

%% file: aw_exp.tex
\section{Harness Scaling in Practice: Two Frontiers and a Hierarchy of Transfer}
\label{sec:experiments}

Harness scaling treats agent capability as a property of both model weights and the execution system that preserves state, reactivates experience, checks progress, and supports repair. Unless otherwise stated, a \textsc{StateM} system comprises the base agent, the generic \textsc{StateM} runtime, and an identified control profile. The results measure this combined system, including the profile's evolved prompts, routing, checks, and practices; they do not isolate the state-machine runtime from the procedural content encoded in the runbook.

Across three axes, we study four empirical regimes: fixed-model lift, frozen transfer within a model family, adapted transfer across providers, and held-out task generalization. The transferable object becomes more abstract as distance grows. Nearby models can share an exact control profile. Across providers, the runtime, runbook structure, applicable practices, and development principles remain reusable, although the profile requires adaptation. Across task distributions, transfer resides in the method for locating and protecting consequential execution boundaries. The question is \emph{what transfers as model and task distance increase}.

\subsection{Evaluation Regimes and Evidence Boundaries}
\label{sec:eval_protocol}

\paragraph{Terminal-Bench 2.1}
Terminal-Bench 2.1 \citep{merrill2026terminal} contains 89 tasks. We run five trials per task, for 445 trials in a full evaluation. The primary metric is trial-level success rate. We also report \emph{five-trial task coverage}, the number of tasks solved at least once across their five trials. This quantity is sometimes reported as Pass@5, but it measures empirical coverage over five observed trials rather than single-run reliability.

Our Terminal-Bench experiments use Codex-based agents with GPT-5.5 xhigh, GPT-5.6 Sol xhigh, GPT-5.6 Luna, and DeepSeek-V4-Flash. Rows labeled \emph{reference} reproduce the corresponding published or public-submission result; rows labeled \emph{our run} report our evaluation. The GPT-5.6 Sol xhigh submission uses \texttt{statem-Codex} agent version \texttt{0.144.1}. Its public submission record contains 445 trials, 439 no-error completions, six \texttt{AgentTimeoutError} trials, 1.178 billion tokens, a standard error of 0.87 percentage points, and a submission-reported model cost of \$1,062.95. It passes all ten automated configuration and static checks in the submission pipeline.

\paragraph{Development, freezing, and adaptation}
We first develop the Terminal-Bench control profile with GPT-5.5. For the GPT-5.6 Sol and Luna evaluations, the versioned profile and its activation policy are frozen: no prompt, state, routing, check catalog, default, or repair policy is revised using target-model evaluation outcomes. Any permitted run-local check is instantiated only from task-visible information under the frozen activation policy, is recorded for that execution, and is not promoted into later trials. The DeepSeek experiment first measures this same frozen profile, then explicitly permits a new adaptation phase before the final evaluation.

\emph{Golden rules} are human-specified constraints on profile development: prefer minimal reusable control, route from visible task semantics rather than task identity, and separate development feedback from frozen evaluation. They govern profile evolution and are distinct from the checks executed inside a run.

In every setting, profile construction may use the visible task specification, workspace artifacts, public documentation, and observable execution feedback. It does not use hidden tests, verifier implementations, public solutions, answer artifacts, task hashes, or manually enumerated task identifiers as routing keys. This boundary prevents direct answer encoding while allowing failure-driven improvement of the harness.

\paragraph{BusinessBench}
BusinessBench \citep{yang2026betterharnessessmallermodels} tests generalization at the task-family level. We construct one profile per treated family, refine it on development instances, and freeze it before the first held-out evaluation. Later refinements using aggregate evaluation feedback are reported separately as post-evaluation diagnostic validation. Each instance has one stochastic trajectory per arm, so family-macro and instance-micro averages answer complementary questions.

\paragraph{Cost accounting}
For DeepSeek, \emph{final-evaluation expenditure} is the realized API charge for producing the reported final-score evidence. \emph{Adaptation expenditure} includes all recorded API charges from the provider-specific profile-development iterations. We report both components and their sum. Public comparator rows list submission-reported model cost separately from the realized API expenditure of our runs. The \$1,062.95 attached to the GPT-5.6 StateM submission is the pipeline's API-equivalent model-cost estimate, not the authors' realized out-of-pocket expenditure under a Codex Pro plan.

\subsection{Same Model, Better Harness: A Model-Generation-Sized Gain}
\label{sec:fixed_model_lift}

\begin{figure*}[t]
    \centering
    \includegraphics[width=0.92\linewidth]{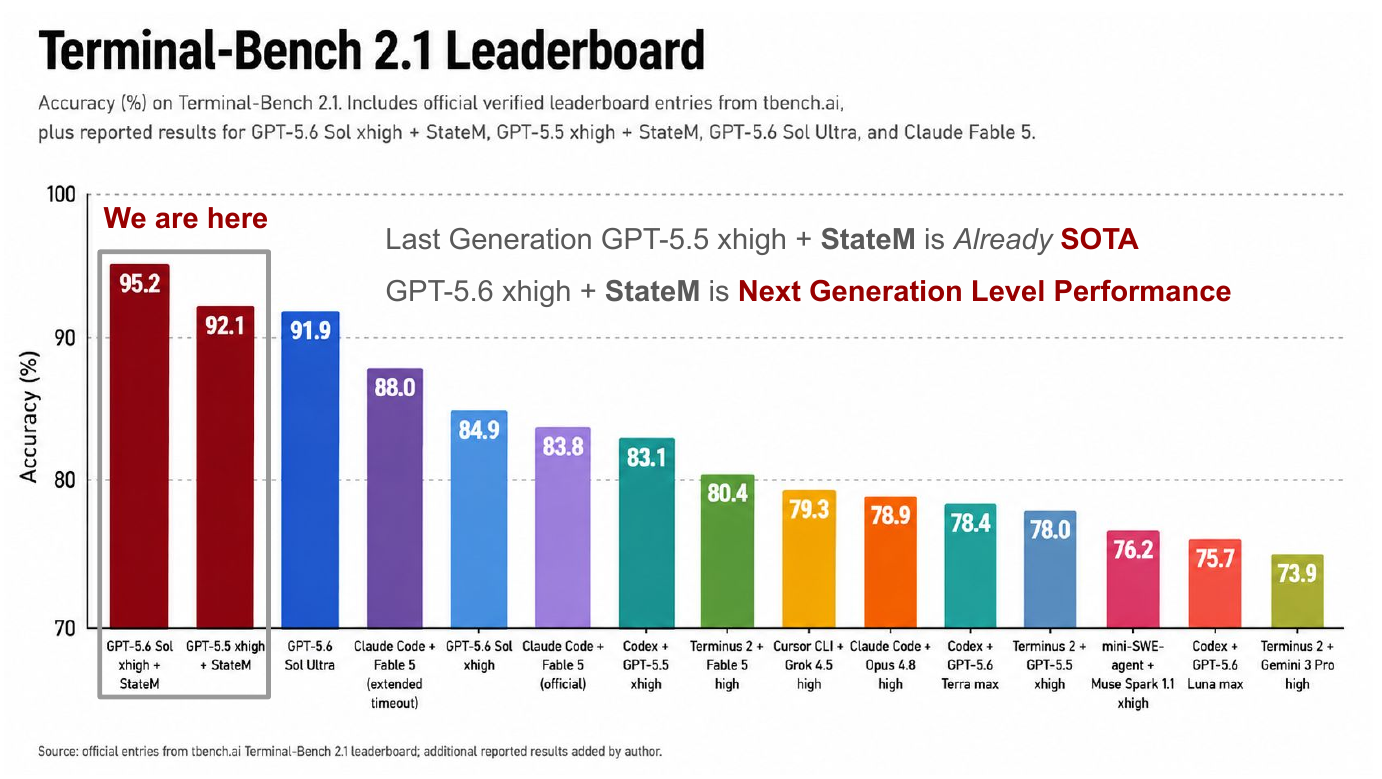}
    \caption{\textbf{A model-generation-sized harness gain on Terminal-Bench 2.1} GPT-5.5 xhigh with StateM records 92.1\%, compared with the 83.1\% GPT-5.5 reference. GPT-5.6 Sol xhigh with the frozen StateM profile records 95.28\% raw accuracy, compared with the 84.9\% Sol xhigh reference. Reference configurations and our runs are visually distinguished; 95.28\% is shown as 95.3\% at one-decimal precision}
    \label{fig:tb21_quality_frontier}
\end{figure*}

Holding model weights fixed isolates the contribution of the harness. With GPT-5.5 xhigh, our StateM system records 92.1\% on Terminal-Bench 2.1, compared with the 83.1\% GPT-5.5 Codex reference. This nine-point difference closes a model-generation-sized gap without changing the underlying model. Across five trials, GPT-5.5 with StateM solves 88 of the 89 tasks at least once, corresponding to 98.9\% five-trial task coverage. Its 92.1\% mean is also numerically above the separately reported 91.9\% GPT-5.6 Sol ultra reference in Fig.~\ref{fig:tb21_quality_frontier}.\footnote{The public GPT-5.5 reference is 83.15\% over 445 trials with Codex agent version \texttt{0.125.0} and \$2,059.19 reported model cost; we round its score to 83.1\%. It is a model-matched public reference rather than a newly rerun, agent-version-matched A/B control. The 91.9\% Sol ultra value is likewise a numeric frontier reference from a different high-compute configuration}

The gain persists with the stronger model. GPT-5.6 Sol xhigh with StateM records
\[
    \frac{424}{445}=95.28\%
\]
raw accuracy, reported as 95.3\% at one decimal, compared with the 84.9\% GPT-5.6 Sol xhigh reference. This is a 10.38-point difference, or 10.4 points at one decimal. The system solves every one of the 89 tasks at least once across five trials. Submission status and adjudication sensitivity are disclosed once in the Introduction; throughout this section, 95.28\% denotes the raw public-submission result.

Under the reference harness, moving from GPT-5.5 to GPT-5.6 Sol changes the score from 83.1\% to 84.9\%, a 1.8-point model-generation shift. The StateM reference differences are +9.0 points on GPT-5.5 and +10.4 points on GPT-5.6 Sol. Here, the execution harness changes completed-task performance far more than the observed generation-to-generation model shift. Model scaling expands what the agent can do; harness scaling determines how reliably that capability becomes a completed job.

\subsection{Transfer Follows Model Distance}
\label{sec:model_transfer}

The fixed-model gains survive model changes at different levels of abstraction, producing a hierarchy rather than a binary transfer outcome.

\paragraph{Frozen transfer within the GPT family}
The Terminal-Bench profile is developed with GPT-5.5 and then frozen before application to GPT-5.6. With no GPT-5.6-specific change to its control logic, the profile accompanies the 84.9\% to 95.28\% Sol xhigh difference reported above and raises GPT-5.6 Luna from 76.7\% to 85.4\%, a gain of 8.7 points. Luna with StateM therefore numerically exceeds the 84.9\% Sol xhigh reference despite using the lower-cost model tier.

Closely related GPT generations and variants share enough recurring execution failures for the same executable control profile to remain useful. The transferred object is concrete: the states, activation policy, routing, checks, and repair structure remain unchanged. ``Frozen transfer'' means zero target-model runbook changes, not zero evaluation cost.

\begin{figure*}[t]
    \centering
    \includegraphics[width=0.84\linewidth]{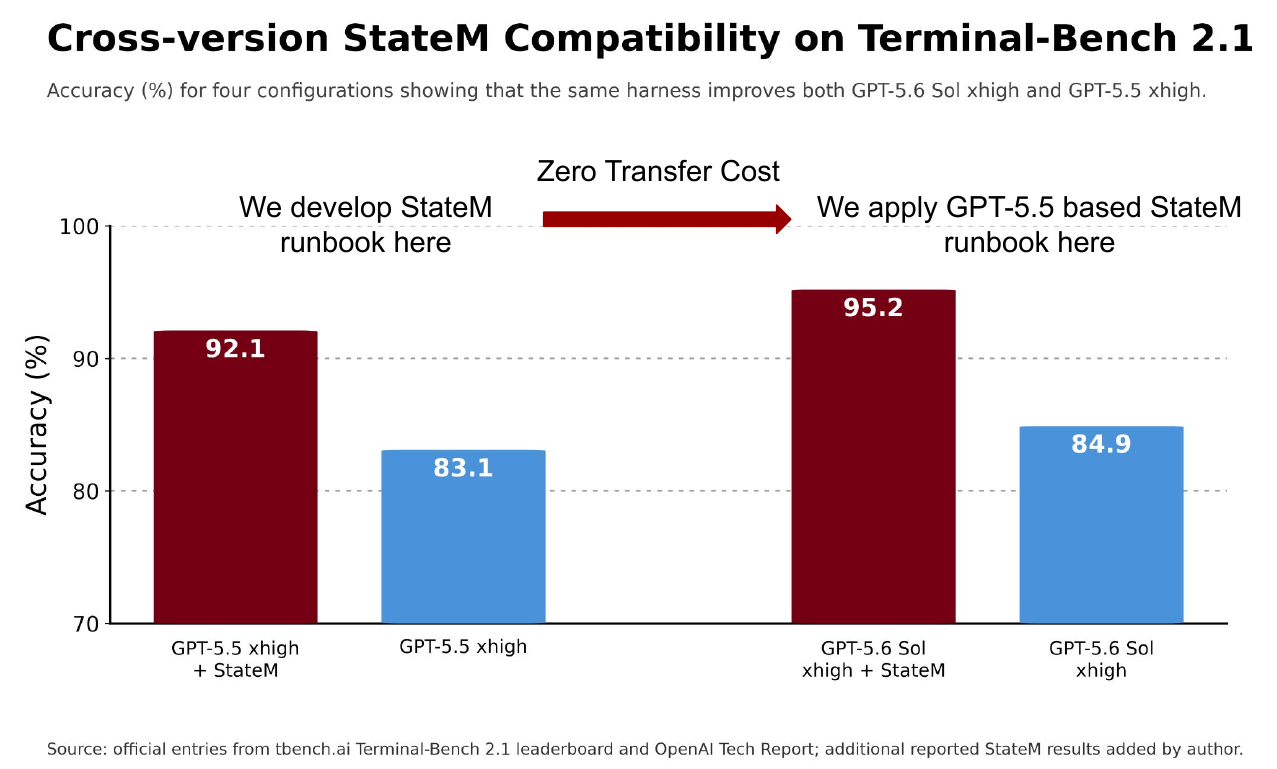}
    \caption{\textbf{Frozen cross-generation transfer with zero target-model runbook changes} The profile developed with GPT-5.5 transfers unchanged to GPT-5.6 Sol xhigh, where the reference difference grows from +9.0 to +10.4 points. The evaluated profile is frozen before any GPT-5.6 outcomes are observed}
    \label{fig:cross_version_transfer}
\end{figure*}

\paragraph{The provider boundary changes the transferable object}
Directly applying the frozen GPT-developed profile to DeepSeek-V4-Flash changes the full-suite score from 82.7\% to 82.0\%. Exact-profile transfer fails across this provider boundary. Because StateM preserves substantial autonomy in the base agent, provider-specific behavior remains consequential even when the task set and control interface are held fixed.

Cross-provider adaptation does not begin from scratch. The generic runtime, high-level runbook structure, routing strategy, already applicable controls, failure-analysis loop, and golden rules remain available. Adapting the concrete practices to DeepSeek's residual failure distribution produces the cost frontier reported in Sec.~\ref{sec:deepseek_frontier}. Transfer follows distance: exact profiles transfer across nearby GPT models, while development principles and the control structure transfer across providers.

\subsection{A \$15 Frontier Run: Moving the Cost--Accuracy Curve}
\label{sec:deepseek_frontier}

DeepSeek-V4-Flash marks the point where exact-profile transfer fails but profile adaptation succeeds. Its 82.7\% baseline changes to 82.0\% under direct frozen transfer of the GPT profile. Starting from the same runtime, runbook structure, applicable controls, and golden rules, provider-specific adaptation raises the full 89-task standard-timeout result to
\[
    \frac{392}{445}=88.09\%,
\]
a gain of 5.39 points over the 82.7\% full-suite baseline. On the disclosed latency-stable common core, which excludes only \texttt{gpt2-codegolf}, the same 392 successes give
\[
    \frac{392}{440}=89.09\%.
\]

For this configuration, the remaining task is limited by the standard timeout rather than absolute solvability: \texttt{gpt2-codegolf} is 0/5 under the standard timeout, but DeepSeek-V4-Flash with StateM solves it in 3 of 5 trials under an extended per-task timeout. Combining those disclosed trials with the 88-task common-core result yields a descriptive full-suite aggregate of
\[
    \frac{392+3}{445}=\frac{395}{445}=88.76\%,
\]
which rounds to 88.8\% and matches the separately reported GPT-5.6 Sol max score at one-decimal precision.\footnote{The standard-timeout full-suite result is 392/445 = 88.09\%; the common-core result is 392/440 = 89.09\%; and the descriptive aggregate is 395/445 = 88.76\% after evaluating only \texttt{gpt2-codegolf} with the extended timeout. Reporting all three numbers keeps the operating condition of each result explicit}

The cost frontier moves even more sharply. The complete DeepSeek final-score evidence costs \$15.20 in realized API charges. Provider-specific adaptation costs \$37.02, bringing all recorded DeepSeek adaptation and final-evaluation expenditure to \$52.22. In contrast, the public GPT-5.6 Sol max Codex submission records \$574.68 in model cost. The DeepSeek evidence uses 2.65\% of that submission's recorded cost, or 1/37.8 as much; its standard-timeout full-suite score is 88.09\%, compared with 83.37\% raw for the matched public GPT artifact. Even the entire \$52.22 adaptation-and-evaluation campaign uses about 1/11 of that amount. The separate OpenAI-reported Sol max score is 88.8\%, which the descriptive DeepSeek aggregate matches to one decimal.\footnote{The 88.8\% Sol max score and the \$574.68 public-submission cost originate from different evaluations. The \$574.68 submission reports 83.37\% raw before adjudication and 76.18\% after adjudication. We therefore use 88.8\% as a score-only reference and plot \$574.68 only at its matched submission score}

Harness transfer changes the economics of deployment. Exact controls did not survive the provider boundary unchanged, but the reusable runbook structure and development principles made adaptation cheap enough to move a lower-cost model sharply upward on the quality--cost frontier. The choice is no longer simply ``buy the strongest model'': one can invest in a harness that turns a cheaper model into a stronger system.

\begin{figure*}[t]
    \centering
    \includegraphics[width=\linewidth]{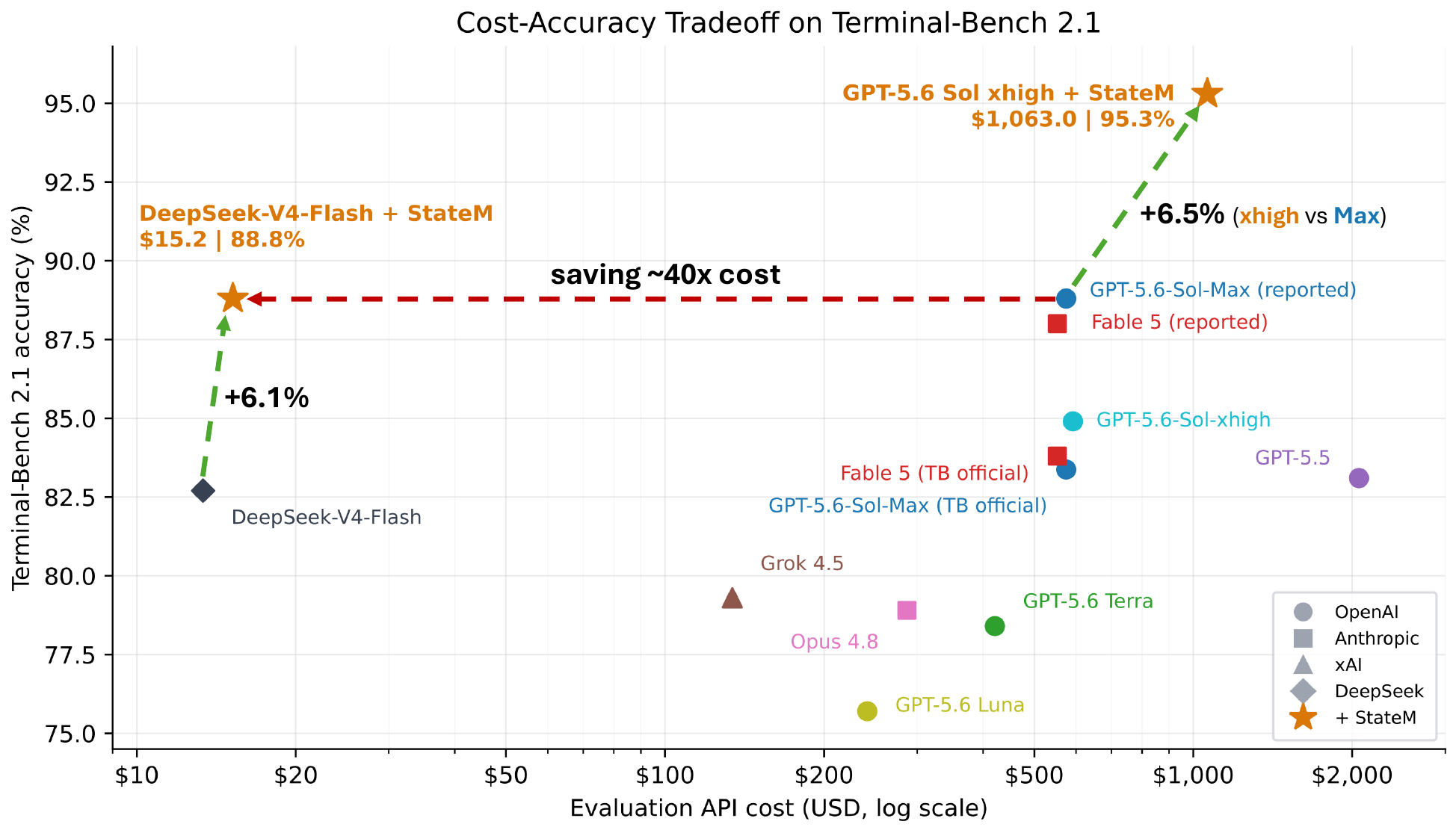}
    \caption{\textbf{StateM moves both the quality and cost frontiers on Terminal-Bench 2.1} The orange points are matched score--cost pairs from our runs: GPT-5.6 Sol xhigh + StateM at 95.28\% raw and \$1,062.95 submission-reported model cost, and DeepSeek-V4-Flash + StateM at 88.76\% descriptive accuracy and \$15.20 realized final-evaluation expenditure. The horizontal 88.8\% reference denotes the separately reported GPT-5.6 Sol max score. The public \$574.68 GPT-5.6 Sol max submission is plotted at its matched 83.37\% raw score. The upper-left direction is better}
    \label{fig:tb21_cost_frontier}
\end{figure*}

\subsection{Task Generalization Follows the Control Boundary}
\label{sec:BusinessBench_generalization}

Terminal-Bench tests one shared profile across heterogeneous tasks. A coherent task family, by contrast, has its own tools, invariants, and completion boundary, and a profile designed for another family may impose irrelevant procedure. We evaluate StateM on BusinessBench~\citep{yang2026betterharnessessmallermodels} using Codex with GPT-5.6 Luna. Family-level profiles are developed on one split and tested on held-out instances from the same family; together, the families test whether the development method remains useful across heterogeneous work.

The benchmark contains 477 eligible instances across seven families. The protocol abstains from applying a StateM workflow to \texttt{attendance-payroll}; all 72 treatment-arm runs in that family receive no StateM intervention and are excluded from efficacy aggregates. The treated set therefore contains 405 instances across six families, or 810 genuine treatment/control executions. GPT-5.6 Luna task agents execute development instances and propose reusable changes from their failure traces; a GPT-5.6 Sol xhigh hyper-agent evaluates family-level generality and compatibility before reconciling the accepted changes. The resulting profile is frozen before the first held-out evaluation.

\paragraph{Frozen transfer improves held-out performance}
On the untouched one-shot held-out split, the equal-family macro average improves from 84.67\% to 85.22\%, a gain of 0.55 points. The instance-weighted micro average improves from 84.44\% to 85.78\%, a gain of 1.34 points. Development performance improves from 86.07\% to 91.71\% (+5.64), and performance over all treated instances improves from 84.76\% to 88.72\% (+3.96).

The largest gains are concentrated in two structurally matched families. Budget Approval improves from 62.91\% to 75.12\% (+12.21), and Machine Operating improves from 90.79\% to 100.00\% (+9.21). Their equal-family held-out macro average rises from 71.91\% to 81.94\%, a 10.04-point gain computed from the unrounded family values. Both families require the workflow to preserve explicit constraints, intermediate state, and completion conditions across standardized tools. Budget Approval benefits from exact-decimal calculation, policy reconciliation, and mandatory-effect closure. Machine Operating benefits from task-derived query planning, data-plane execution, interval coverage, and durable publication.

\begin{table*}[t]
\centering
\small
\setlength{\tabcolsep}{7pt}
\renewcommand{\arraystretch}{1.18}
\begin{tabular}{l c c c}
\toprule
Scope / task family & Codex CLI & StateM--Codex & $\Delta$ \\
\midrule
\multicolumn{4}{l}{\textit{Frozen one-shot aggregate results}} \\
Held-out, family macro & 84.67 & \textbf{85.22} & $+0.55$ \\
Held-out, instance micro & 84.44 & \textbf{85.78} & $+1.34$ \\
Development & 86.07 & \textbf{91.71} & $+5.64$ \\
All StateM-treated & 84.76 & \textbf{88.72} & $+3.96$ \\
\midrule
\multicolumn{4}{l}{\textit{Exploratory mechanism-matched held-out subgroup}} \\
Budget Approval + Machine Operating, family macro & 71.91 & \textbf{81.94} & $+10.04^{\dagger}$ \\
\midrule
\multicolumn{4}{l}{\textit{Frozen Round-1 family aggregates}} \\
\texttt{budget-approval} & 62.91 & \textbf{75.12} & $+12.21$ \\
\texttt{machine-operating} & 90.79 & \textbf{100.00} & $+9.21$ \\
\texttt{refactorbench} & \textbf{80.56} & 77.78 & $-2.78$ \\
\texttt{webarena} & 88.00 & \textbf{92.00} & $+4.00$ \\
\texttt{webtest} & 98.25 & \textbf{98.75} & $+0.50$ \\
\texttt{woocommerce-stock} & \textbf{92.59} & 88.89 & $-3.70$ \\
\midrule
\multicolumn{4}{l}{\textit{Abstention control, excluded from StateM aggregates}} \\
\texttt{attendance-payroll} & 88.43 & \texttt{abstained} & \texttt{n/a} \\
\bottomrule
\end{tabular}
\caption{\textbf{Frozen one-shot BusinessBench results with Codex + GPT-5.6 Luna} The first four rows summarize the six treated families; family rows report the corresponding Round-1 aggregates. Attendance is excluded because no StateM workflow is applied. $^{\dagger}$The 10.04-point subgroup delta is computed from unrounded source values; the displayed values differ by 10.03 after independent rounding. The 80.56 RefactorBench Round-1 baseline and the later 80.56 StateM development score in Table~\ref{tab:BusinessBench_diagnostic} belong to different runs and splits}
\label{tab:BusinessBench_frozen}
\end{table*}

\paragraph{Negative transfer identifies the wrong boundary}
The first frozen profiles do not improve every family. RefactorBench changes from 80.56\% to 77.78\% ($-2.78$), and WooCommerce Stock changes from 92.59\% to 88.89\% ($-3.70$). In both cases, the problem is not simply too little control. The control is attached to the wrong execution boundary.

RefactorBench initially overemphasizes minimality and backward compatibility but fails to close explicit code-migration obligations. A later, thinner profile tracks visible obligations, verifies required signature and call-shape changes, and performs one repository-wide stale-reference closure. In a post-evaluation matched rerun, the overall score changes from 76.39\% to 79.17\%. On the development split, the revised StateM arm improves from 75.00\% to 80.56\%; on the reused held-out split, it remains 77.78\%. The rerun locates the original negative transfer in profile mis-specification: the useful intervention is less general compatibility procedure and more direct obligation closure.

WooCommerce exhibits a different boundary error. Its first profile contains substantial procedure but fails to preserve the cross-system invariants that govern inventory entities, destination-specific deduplication, irreversible email actions, and receipts. After replacing generic control with these invariants, a post-evaluation matched rerun improves overall performance from 86.42\% to 90.12\%, including an improvement from 85.37\% to 90.24\% on the reused held-out split.

WebArena is dominated by retrieval and analysis. Query modifiers, page state, sorting, and pagination matter more than a mutation-oriented workflow. Its frozen evaluation contains only three non-tied A/B outcomes across 25 instances: development has 2 wins, 10 ties, and 0 losses; held-out has 0 wins, 12 ties, and 1 loss. A proposed minimal query-state profile has not yet been evaluated due to model capacity constraint. WebTest is nearly saturated: later matched reruns reach 100.0\% on development and 99.5\% on the reused held-out split for both arms, or 99.75\% overall.

\begin{table*}[t]
\centering
\small
\setlength{\tabcolsep}{6pt}
\renewcommand{\arraystretch}{1.17}
\begin{tabular}{l c c l}
\toprule
Post-evaluation validation scope & Codex / prior arm & Refined StateM & Interpretation \\
\midrule
Held-out, instance micro & \texttt{n/a} & 86.67 & Selectively refined aggregate \\
Development & \texttt{n/a} & 92.21 & Selectively refined aggregate \\
All StateM-treated & \texttt{n/a} & 89.42 & Selectively refined aggregate \\
\midrule
\texttt{refactorbench}, overall & 76.39 & \textbf{79.17} & Matched rerun \\
\texttt{refactorbench}, development & 75.00 & \textbf{80.56} & Obligation-and-closure profile \\
\texttt{refactorbench}, reused held-out & 77.78 & 77.78 & No held-out change \\
\texttt{webtest}, overall & 99.75 & 99.75 & Saturated \\
\texttt{webtest}, development & 100.0 & 100.0 & Saturated \\
\texttt{webtest}, reused held-out & 99.5 & 99.5 & Saturated \\
\texttt{woocommerce-stock}, overall & 86.42 & \textbf{90.12} & Invariant-matched profile \\
\texttt{woocommerce-stock}, reused held-out & 85.37 & \textbf{90.24} & Invariant-matched profile \\
\bottomrule
\end{tabular}
\caption{\textbf{Post-evaluation diagnostic validation after selective profile refinement} These results use aggregate evaluation feedback and are separated from the untouched one-shot evidence in Table~\ref{tab:BusinessBench_frozen}. The first three rows are refined StateM aggregate scores; the remaining rows are matched reruns or split-level diagnostics}
\label{tab:BusinessBench_diagnostic}
\end{table*}

Across the seven families, \textbf{harness generalization follows mechanism match, not task diversity}. RefactorBench requires code-contract and stale-reference closure; WebArena requires query and page-state control; WebTest primarily requires a tool-execution boundary; WooCommerce requires entity, destination, and irreversible-side-effect invariants; and Attendance shows that abstention can be the correct control decision. A strong profile is not a maximal universal workflow. It binds the minimum invariants likely to drift and verifies them where a violation becomes consequential. Appendix Table~\ref{tab:generalized-statem-controls} lists the generalized controls distilled from these families.

\subsection{Failure-Boundary Evidence: Where StateM Intervenes}
\label{sec:failure_boundary_evidence}

Control-signal dilution and mutable-state ambiguity create the operational pressures that make long runs difficult. Epistemic, procedural-compliance, and procedural-memory gaps locate the resulting failures. StateM supplies a distinct control point for each gap.

For an \emph{epistemic gap}, a state-local \texttt{in\_hook} can reactivate relevant domain knowledge, tool guidance, or task context at the phase where it becomes useful. For a \emph{procedural-compliance gap}, states, checks, transition guards, and evidence requirements can block a configured incomplete handoff and keep the run in a repairable state. For a \emph{procedural-memory gap}, a selected lesson from one execution can become a versioned prompt, check, activation rule, or recovery path for later independent runs.

Table~\ref{tab:task-level-improvements} reports representative GPT-5.5 task-level evidence. The largest gains cluster around consequential boundaries: service readiness before handoff, contract satisfaction before committing a transformation, preservation before destructive operations, and evidence closure before completion. The table associates each gain with the StateM control active in that task; the association is not a component ablation.

\begin{table*}[t]
\centering
\small
\setlength{\tabcolsep}{6pt}
\renewcommand{\arraystretch}{1.20}
\begin{tabular}{l c c c >{\raggedright\arraybackslash}p{6.0cm}}
\toprule
Task & Codex CLI & StateM--Codex & $\Delta$ & Associated StateM control \\
\midrule
\texttt{configure-git-webserver} & 0/5 & 5/5 & \textbf{$+5$} & Service/deploy state and consumer-facing verification \\
\texttt{dna-insert} & 0/5 & 5/5 & \textbf{$+5$} & State-local biological and primer contract checks \\
\texttt{dna-assembly} & 1/5 & 5/5 & \textbf{$+4$} & Transition gate for primer, Tm, and assembly invariants \\
\texttt{filter-js-from-html} & 0/5 & 4/5 & \textbf{$+4$} & HTML/script extraction boundary checks and negative controls \\
\texttt{db-wal-recovery} & 2/5 & 5/5 & $+3$ & Preflight preservation before destructive consumers \\
\texttt{sanitize-git-repo} & 4/5 & 5/5 & $+1$ & Modification-range verification \\
\texttt{protein-assembly} & 2/5 & 5/5 & $+3$ & Constraint checklist and self-review \\
\texttt{pypi-server} & 3/5 & 5/5 & $+2$ & Service-lifetime and readiness gate \\
\texttt{install-windows-3.11} & 3/5 & 5/5 & $+2$ & VM setup and lifecycle completion \\
\texttt{pytorch-model-recovery} & 3/5 & 5/5 & $+2$ & Task-visible model evidence \\
\texttt{qemu-alpine-ssh} & 2/5 & 4/5 & $+2$ & Service readiness and repeated consumer check \\
\texttt{extract-moves-from-video} & 0/5 & 2/5 & $+2$ & Candidate-first bounded refinement \\
\bottomrule
\end{tabular}
\caption{\textbf{Representative Terminal-Bench 2.1 task-level improvements for GPT-5.5} Each result is the number of successful trials out of five for one task. The listed controls were proposed and implemented by GPT-5.5 Codex under the human-specified high-level design and golden rules. They are selected from visible task semantics and workspace evidence, without using task identifiers, task names, hashes, hidden tests, verifier implementations, or answer artifacts as activation keys}
\label{tab:task-level-improvements}
\end{table*}

The \texttt{configure-git-webserver} task isolates this distinction. The baseline agent can configure Git, SSH, hooks, and an HTTP server, yet scores 0/5 because it does not reliably preserve and validate the required end-to-end live state. With StateM, final handoff is gated on fresh consumer-facing evidence: the agent must materialize a clone--commit--push--curl path before leaving verification. If verification perturbs the environment, the run remains in a repairable state until final-state consistency is restored. StateM adds no new component-level capability in this case. It composes, checks, and closes capabilities already available to the model before the consequential handoff boundary. The result moves from 0/5 to 5/5.

Epistemic hooks determine which knowledge is active at a state; checked transitions determine whether a known procedure is completed; and versioned practices determine whether a lesson survives into future runs. StateM preserves the base agent's freedom inside each phase while adding the minimum persistent control where a recoverable error would otherwise become final.

\subsection{Harness Learning as Selective Procedural Memory}
\label{sec:procedural_memory}

The runbook is both a runtime control surface and a selective memory for lessons that persist across independent executions. We call a reusable experience-derived intervention a \emph{practice}. A practice may be a state-local instruction, a check, a constraint, an activation condition, or a verification action triggered when visible evidence indicates elevated downstream failure risk. Encoding the lesson in a versioned profile makes it inspectable and executable; a later run no longer has to rediscover it from a postmortem paragraph.

\paragraph{From one failure to one candidate practice}
At the single-task level, learning follows a concrete loop: execute the current profile, observe failure or verifier feedback, attribute a candidate cause, abstract a reusable lesson, encode a profile change, and validate it again. Because the runbook is agent-native, the executor and maintainer can be the same agent. The agent can change a state-local prompt, add an evidence check, revise a transition, or introduce a repair path without reimplementing the entire reasoning process as a collection of controller-owned graph nodes.

\paragraph{From task traces to a family profile}
Multi-task settings make the abstraction step decisive. A lesson from one development instance should not automatically become a rule for the family. In BusinessBench, GPT-5.6 Luna task agents execute development instances and propose reusable changes. A stronger hyper-agent evaluates their family-level generality and compatibility before reconciling them into a shared family profile. The frozen one-shot evaluation tests whether that abstraction survives on held-out tasks.

\paragraph{One broad benchmark requires sparse routing}
Terminal-Bench requires one shared profile to cover 89 heterogeneous tasks, while the accumulated trajectories exceed what a hyper-agent can review in one context. Repairing tasks one at a time causes control drift: a local rule can add unnecessary procedure or damage a different task. We use a thin shared runbook, generalizable routing from visible task semantics, and late state-local checks activated only when the current task and workspace evidence make them relevant. For epistemic gaps such as statistical reasoning in \texttt{raman-fitting}, the corresponding \texttt{in\_hook} can expose general public domain knowledge gathered by the hyper-agent. Human input supplies the high-level architecture and golden rules; GPT-5.5 Codex produces most of the Terminal-Bench harness implementation and performs much of the low-level trace analysis and proposal generation.

The BusinessBench iterations show why this memory must be selective. More remembered procedure is not better memory. RefactorBench improves when repeated compatibility review is replaced by a smaller obligation-and-closure practice. WooCommerce improves when generic procedure is replaced by the actual cross-system invariants. WebArena appears to require a still lighter query-state boundary. Attendance shows that the correct policy can be not to introduce a StateM workflow at all. Effective procedural memory requires addition, revision, and removal.

\paragraph{What the harness should not remember}
Failure-driven optimization can preserve the wrong lesson. The Terminal-Bench video tasks do not fully specify target-frame precision. During development, the hyper-agent introduces default precision values after observing benchmark behavior. These defaults may be operationally useful, but they also show how an ambiguous specification can be resolved post hoc and then stored as if it were a universal practice.

Evaluator feedback enters memory through the same route. In the DNA insertion tasks, the verifier selects the left-most valid insertion boundary even though that convention is absent from the visible task description. Repeated feedback can lead the profile to reproduce that convention without ever reading verifier code. The resulting behavior agrees with the evaluator, but its semantics come from the evaluator rather than the stated task contract.

The first BusinessBench profiles reveal a third failure mode: even a valid failure can be abstracted incorrectly. RefactorBench receives excessive compatibility procedure, retrieval-oriented WebArena receives a workflow that is too heavy, and early WooCommerce control misses the invariants that connect its systems. Stronger-agent or human review matters at this level. Failure traces provide experience; deciding which abstraction deserves to become durable control is itself a difficult reasoning task.

The design rule is direct: \textbf{experience must be filtered before it becomes memory}. Harness scaling is an abstraction problem, not rule accumulation. The objective is to remember the consequential boundary, not every failure trace.\footnote{BusinessBench family sizes are unequal and each instance has one stochastic trajectory per arm. Only the first frozen held-out evaluation is untouched; later results are explicitly post-evaluation validation. StateM can organize available knowledge and execution control, but it cannot remove limits imposed by the base model, tools, environment, or execution budget}

\subsection{Operational Endurance: A 22-Hour Development Run}
\label{sec:operational_endurance}

StateM was built for an operational requirement beyond benchmark accuracy: a long-running agent must preserve its place and obligations as the interaction grows. In one Terminal-Bench profile-development run, the hyper-agent continued for 22 hours across long interaction history, context refresh or compaction, and stop-hook continuation. Throughout the run, the durable StateM record provided the current phase, transition history, unresolved obligations, and recovery anchor. The observed slowdown came from accumulated terminal output that remained unfolded in the interface, rather than loss of the StateM control state.

The runtime sustains a day-scale development loop while keeping procedural state external to the changing model context; the result does not imply unbounded execution. Together with the benchmark results, this completes the transfer hierarchy: exact runbooks transfer across nearby models, development principles transfer across providers, and control-boundary abstractions transfer across tasks.